\documentclass{article}

\PassOptionsToPackage{numbers, compress}{natbib}

\usepackage[final]{neurips_2025}

\usepackage[utf8]{inputenc} 
\usepackage[T1]{fontenc}    
\usepackage{hyperref}       
\usepackage{url}            
\usepackage{booktabs}       
\usepackage{amsfonts}       
\usepackage{nicefrac}       
\usepackage{microtype}      
\usepackage{xcolor}         
\usepackage{graphicx}
\usepackage{subfigure}
\usepackage{amsmath}
\usepackage{amssymb}
\usepackage{mathtools}
\usepackage{amsthm}
\usepackage{csquotes}
\usepackage{xspace}
\usepackage{algorithm}
\usepackage{algorithmic}
\usepackage{multirow}
\usepackage{soul} 
\usepackage{color}
\usepackage{wrapfig}
\usepackage[capitalize]{cleveref}
 
\theoremstyle{plain}

\theoremstyle{definition}

\theoremstyle{remark}

\usepackage[separate-uncertainty=true]{siunitx}
\usepackage[textsize=tiny]{todonotes}

\newcommand{\ours}{REP\xspace}
\newcommand{\atom}{AToM\xspace}
\newcommand{\ald}{ALD\xspace}

\newcommand{\parlabel}[1]{\vspace{0.2em}\noindent\textbf{#1}}

\makeatletter
\DeclareRobustCommand\onedot{\futurelet\@let@token\@onedot}
\def\@onedot{\ifx\@let@token.\else.\null\fi\xspace}
\def\eg{{e.g}\onedot}

\def\ie{{i.e}\onedot}

\makeatother
\usepackage[neverdecrease]{paralist}
\usepackage{enumitem}

\title{REP: Resource-Efficient Prompting for Rehearsal-Free Continual Learning}

\author{
Sungho~Jeon, Xinyue~Ma, Kwang~In~Kim, Myeongjae~Jeon\\
POSTECH \\
\texttt{\{sunghojeon,xinyuema,kimkin,mj.jeon\}@postech.ac.kr} \\
}

\begin{document}

\maketitle

\begin{abstract}
Recent rehearsal-free continual learning (CL) methods guided by prompts achieve strong performance on vision tasks with non-stationary data but remain resource-intensive, hindering real-world edge deployment. We introduce resource-efficient prompting (\ours), which improves the computational and memory efficiency of prompt-based rehearsal-free continual learning methods while minimizing accuracy trade-offs. Our approach employs swift prompt selection to refine input data using a carefully provisioned model and introduces adaptive token merging (\atom) and adaptive layer dropping (\ald) for efficient prompt updates. \atom and \ald selectively skip data and model layers while preserving task-specific features during the learning of new tasks. Extensive experiments on multiple image classification datasets demonstrate \ours's superior resource efficiency over state-of-the-art rehearsal-free CL methods.
\end{abstract}

\section{Introduction}
\label{sec:intro}

Continual learning (CL) trains neural network models on multiple sequential tasks, where each task may include data distributions that diverge from previously encountered data. A crucial challenge for any CL algorithm is to effectively address \emph{catastrophic forgetting}~\cite{mccloskeyC89}. Severe forgetting occurs when a model rapidly loses previously learned knowledge while adapting to new tasks, significantly affecting the model's reliability and accuracy on earlier tasks.

Moreover, many of today’s AI services are designed for on-device scenarios to securely learn tasks locally~\cite{kwon2021exploring, hayes2022online, e-domainil}.
In on-device CL, improving \emph{computational efficiency} is crucial, as it directly reduces energy usage and enhances the durability of edge devices. Meanwhile, since device memory capacity often acts as a hard constraint, CL tasks that exhaust all available memory can cause system crashes due to out-of-memory errors~\cite{sage}. Given typically limited memory sizes (1--8GB), achieving high \emph{memory efficiency} is essential to enable diverse on-device deployment scenarios in practice.

\begin{figure*}[t]
    \vskip 0.2in
    \begin{center}
    \centerline{\includegraphics[width=\textwidth]{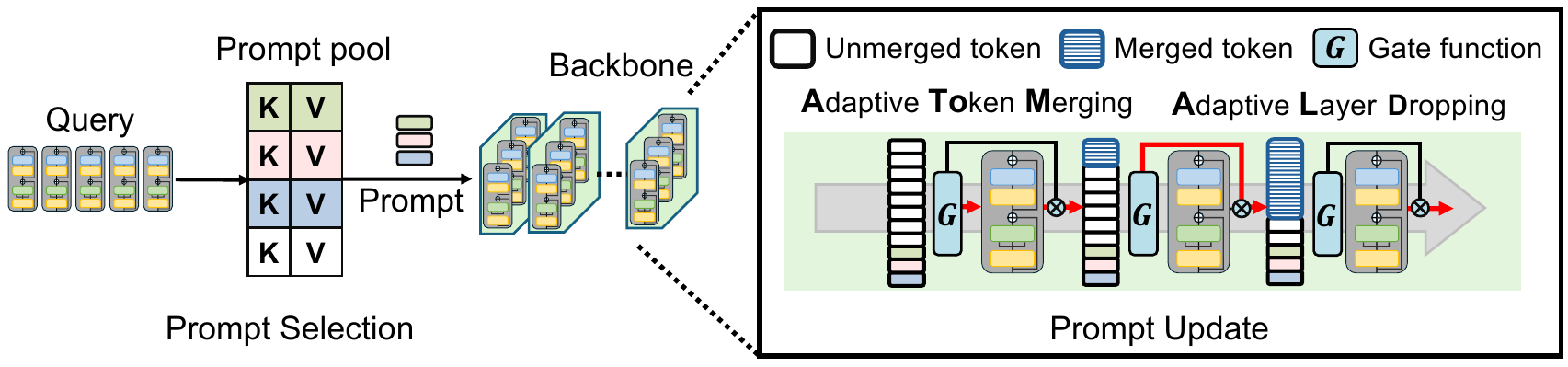}}
    \caption{Overview of the proposed resource-efficient prompting (\ours) algorithm for rehearsal-free CL. \ours calculates
    query features from input samples using a lightweight surrogate model (\eg, ViT-Ti) and random projections
    to swiftly extract prompts from the prompt pool. These prompts are then
    inserted into a main backbone model (\eg, ViT-L) for training, which
    prioritizes model accuracy.}
    \label{fig:repoverview}
    \vskip -0.2in
    \end{center}
\end{figure*}

In this paper, we propose \ours(\cref{fig:repoverview}), a framework for \emph{resource efficiency} in prompt-based rehearsal-free CL methods on frozen, pre-trained vision transformers (ViTs)~\cite{dualp, l2p, sprompts, codap, dap, ovor, convp, hidep}. Prompts are a small set of parameters that progressively learn incoming tasks to combat forgetting. Updates to these compact prompts incur minimum data writes without significantly harming the lifespan of device storage, which typically sustains up to 10K of writes per location. This makes prompt-based rehearsal-free methods well-suited for on-device continual learning.

\ours is built upon our analysis of cost-accuracy trade-offs throughout the end-to-end learning process, with two key design insights. (1) The \emph{prompt selection} stage, which constructs a prompt subset to augment input or intermediate data is highly amenable to numerous promising options with fast approximations. (2) In contrast, the \emph{prompt update} stage, which involves forward-backward passes over the backbone model, presents a range of optimizations with vastly different cost-accuracy trade-offs. Based on these insights, we develop three complementary techniques for both stages that trade negligible accuracy drop for high resource efficiency: (1) \emph{random projection-based lightweight surrogate model} for prompt selection, and (2) \emph{adaptive token merging} (\atom) and (3) \emph{adaptive layer dropping} (\ald) for prompt update. Notably, \atom and \ald \emph{non-uniformly} skip
parts of the data and model layers to reduce training costs while preserving critical task-specific knowledge. These approaches are inspired by our analysis (\cref{fig:vitanalysis}) and prior works~\cite{dualp, vitnature}
demonstrating that shallower layers in pre-trained ViTs are more sensitive to new tasks than deeper layers.

Experimental results demonstrate that \ours considerably reduces training time (up to 51\%) and memory usage (up to 41\%), with only a marginal accuracy drop, when applied to seven state-of-the-art prompt-based CL methods across three backbone models of varying sizes. As the proposed techniques are complementary in optimizing resource efficiency, we further evaluate them by integrating them into two distinct ViT-based, rehearsal-free CL methods that do not rely on prompting. In these settings, \ours achieves a 37--48\% reduction in training time and up to a 48\% decrease in memory usage, confirming that its efficacy can extend well beyond prompt-based CL scenarios.

\section{Related work}
\label{sec:related}

\parlabel{Class-incremental learning.\;}
Our work focuses on a class-incremental learning (CIL) setting, where each task introduces new classes within the same domain, but task IDs are unavailable during inference~\cite{gepperthH16}. In CNN-based approaches, \emph{rehearsal-based} methods that replay stored samples remain dominant due to their strong performance~\cite{rainbow, castro2018eccv, rwalk, gdumb, icarl}. Recent methods such as BudgetCL~\cite{budgetcl} and CarM~\cite{carm} adopt data-driven strategies that outperform optimization-based methods under computational constraints. MEMO~\cite{memo} trades stored samples for task-specific layers when beneficial. We compare our prompt-based approach with these methods and show superior performance under resource constraints.

\parlabel{Continual learning for the edge.\;}
Recent efforts in on-device learning have emphasized memory and energy efficiency, primarily outside the CL domain~\cite{sage, melon, e2train}. A few studies have extended CL to edge devices: \citet{hayes2022online} investigated online CL with CNNs for embedded systems; \citet{kwon2021exploring} analyzed rehearsal vs. regularization methods in terms of storage, compute, and accuracy trade-offs; and \citet{miro} introduced Miro, a platform that dynamically configures CNN-based CL to reduce energy consumption within memory constraints. However, none of these works address the challenge of resource-efficient vision transformers (ViTs) for on-device CL, which is the primary focus of our work.

\parlabel{Prompting for continual learning.\;} 
Prompting, which provides explicit instructions or queries to the model, has proven effective for ViTs in CL by enabling adaptation to new tasks while preserving prior knowledge~\cite{ovor, dap, convp, codap, hidep, sprompts, dualp, l2p}. L2P~\cite{l2p} and DualPrompt~\cite{dualp} retrieve task-relevant prompts from a shared pool using prompt tuning and prefix tuning, respectively. CODA-Prompt~\cite{codap} improves prompting capacity with attention-conditioned prompts, while OVOR~\cite{ovor} uses a single prompt to accelerate selection. HiDe-Prompt~\cite{hidep} extends supervised prompt-based CL to self-supervised settings through hierarchical optimization. ConvPrompt~\cite{convp} employs convolution-based, layer-wise prompts to improve knowledge transfer with negligible overhead. LAE~\cite{lae} interprets prompts as a form of parameter-efficient fine-tuning, incorporating modules such as adapters or LoRA. As shown in~\cref{sec:experiments}, our techniques can be applied to these methods to enhance resource efficiency without significant accuracy loss.

\section{\ours: Resource-Efficient Prompting}
\label{sec:design}

Prompt-based rehearsal-free CL achieves strong performance by mitigating forgetting without relying on replay. However, it remains resource-intensive during the prompt selection and update stages. In particular, prompt updates require full forward and backward passes, which are costly for large backbones, limiting their applicability on resource-constrained devices. 

\subsection{Motivation: challenges in prompt selection and update}

\textbf{Prompt selection} operates a neural network $f_\text{query}$ to compute a query feature $q(x_i^j) \in \mathbb{R}^D$ for a given input $x_i^j$ from task $j$, It then selects the prompt $p^*$ from the prompt pool $P$ that maximizes cosine similarity with the query feature:
\begin{align}
p^* &= \underset{p_k \in P}{\mathrm{argmax}}\ \frac{\langle q(x_i^j), p_k\rangle}{\|q(x_i^j)\|\|p_k\|}. \label{eq:selection_query_a}
\end{align} 
In practice, $f_\text{query}$ is often the same as the main backbone $f_\text{update}$, and each query adds an extra forward pass per sample. For large models such as ViT-L, this results in up to 28\% increase in computation time (see~\cref{tab:component_ablation}).

\vspace{1mm}
\noindent\textbf{Prompt update} refines learnable parameters for effective training and adaptation by combining the classification loss $L_\text{class}$, a prompt-specific loss $L_{\text{prompt}}$, and an auxiliary loss $L_{\text{aux}}$ during task $j$:
\begin{align}
L
=
L_{\text{class}}\bigl(f_\text{update}(x_i^j),\,y_i^j\bigr)
\;+\;
\epsilon_1\,L_{\text{prompt}}\bigl(p^*,\,q(x_i^j)\bigr)
\;+\;
\epsilon_2\,L_{\text{aux}},
\label{eq:loss_function}
\end{align}
where $f_{\text{update}}$ denotes the main backbone, and $\epsilon_1$ and $\epsilon_2$ control the influence of the prompt-specific and auxiliary losses. Different methods instantiate this loss differently. For example, \cite{dualp, l2p} omit the auxiliary loss by setting $\epsilon_1=1, \epsilon_2=0$, whereas \cite{codap, hidep} tune both coefficients.

Although most backbone layers remain frozen, each mini-batch still incurs full forward and backward passes through the network, requiring storage of intermediate activations. As a result, large backbones such as ViT-L exhibit substantial memory usage and prolonged training time (see~\cref{tab:comparison}). 

\subsection{Proposed solutions: \ours}
\label{subsec:systemefficiency}

REP is built on two core insights:
\setdefaultleftmargin{0.5cm}{0cm}{}{}{}{}
\begin{compactenum}
\item The prompt selection stage can tolerate approximations: high retrieval quality can be achieved without relying on the full-capacity backbone.
\item Not all layers in a frozen backbone contribute equally to adaptation, allowing selective computation during prompt updates.
\end{compactenum}
To realize these insights, REP introduces three complementary techniques: a lightweight surrogate model for efficient prompt selection, and adaptive token merging (AToM) and adaptive layer dropping (ALD) for efficient prompt updates.

\subsubsection{Guiding prompt selection with a lightweight surrogate model}
\label{sec:prompt_downsizing}
To reduce the computational cost of prompt selection, we use a lightweight surrogate model $f_\text{efficient}$ with reduced depth and width. Instead of reusing a large backbone such as ViT-L, we adopt a compact pre-trained ViT-Ti model, which effectively captures essential representations. Given an input $x_i^j$, $f_\text{efficient}$ produces a low-dimensional query feature $q_{\text{efficient}}(x_i^j) \in \mathbb{R}^d$ (where $d\le D$). We then apply a fixed, non-trainable random projection $\phi$~\cite{ranpac} to map this feature back to the original $D$-dimensional space for prompt selection:

\begin{align}
p_{\text{efficient}}^* &= \underset{p_k \in P}{\mathrm{argmax}}\ \frac{\langle \phi(q_{\text{efficient}}(x_i^j)), p_k \rangle}{\| \phi(q_{\text{efficient}}(x_i^j)) \| \| p_k \|}. \label{eq:selection_query_b}
\end{align}
Empirically, this strategy preserves around 97\% of the representational similarity to a ViT-L-based query model, as measured by centered kernel alignment~\cite{cka}. Despite residing in a low-dimensional space, $p^*_\text{efficient}$ effectively approximates the large model's representations, substantially reducing computational cost (see~\cref{tab:component_ablation}).

\subsubsection{Analysis of the frozen backbone during prompt update}

\begin{figure}[ht]
    \centering
    \includegraphics[width=\textwidth]{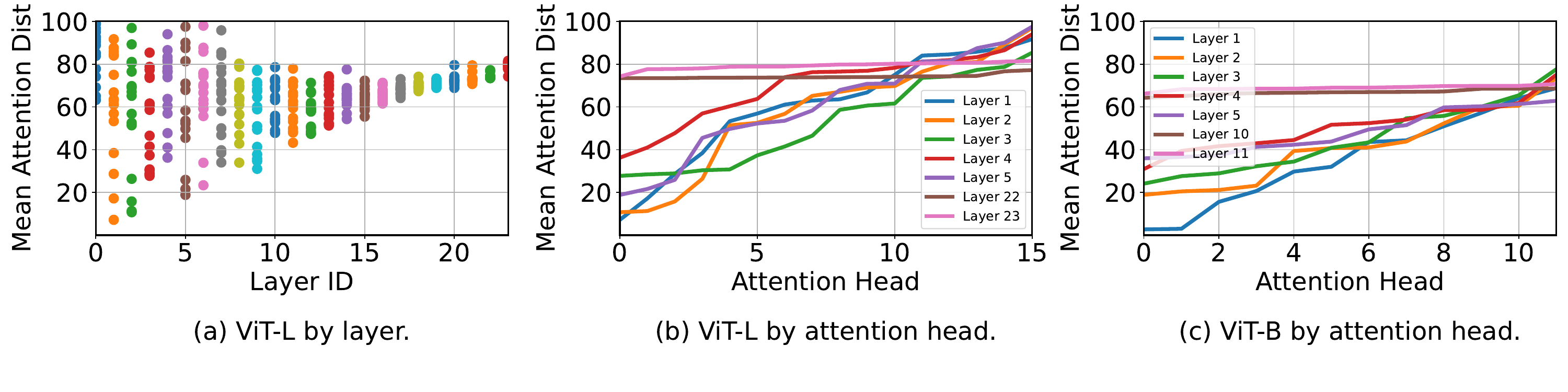}
    \vspace{-5mm}
    \caption{Mean attention distances for frozen blocks along \textbf{(a)} layers
    and \textbf{(b/c)} attention heads. We run the first task of Split ImageNet-R.
    \textbf{(a/b)} L2P with ViT-L, and \textbf{(c)} DualPrompt with ViT-B.}
    \label{fig:vitanalysis}
\end{figure}

A key insight of \ours is that not all frozen blocks contribute equally to the final loss~\(L\). To validate this, we analyze the feature representations of pre-trained transformer blocks with prompts using \emph{mean attention distance}, a metric previously used to study layer-wise representations in ViT models during pre-training~\cite{vit_original, vitnature}.

For a given attention head within a frozen block, let \((x_q, y_q)\) denote the position of a query patch, and \((x_i,y_i)\) the positions of the patches the head attends to, with corresponding attention weights \(a_i\). The
distance \(d_i\) between patches---typically defined as Euclidean or pixel distance---is given by \(d_i = (x_i - x_q)^2 + (y_i - y_q)^2\). We then compute the weighted mean distance as \(\frac{\sum_i a_i \cdot d_i}{\sum_i
a_i}\). Following~\cite{vit_original,vitnature}, a smaller distance indicates attention focused on a narrow region (local information), while a larger distance suggests more distributed attention across the input (global
information). We measure the average mean attention distance of L2P with the ViT-L backbone along two dimensions: layer ID and attention head, during adaptation to a new task in Split ImageNet-R (10 tasks).

As shown in~\cref{fig:vitanalysis}(a), shallower layers exhibit a wider range of attention distances and tend to focus on both localized and global regions. In contrast, deeper layers consistently produce higher values, reflecting a shift toward more global contextual information. This pattern aligns well with prior findings in~\cite{vit_original,vitnature}. A closer look at a few selected layers in~\cref{fig:vitanalysis}(b) further
reveals that attention distances across heads vary more widely in shallow layers than in deeper ones, confirming that shallower layers capture more important representations of the input.

A similar trend is observed when using DualPrompt with ViT-B in~\cref{fig:vitanalysis}(c). Unlike L2P, DualPrompt attaches prompts to the self attention layer of multiple transformer blocks, rather than only in the input sequence. This suggests that the feature representations of the pre-trained model, which balance local and global information, may not be greatly affected by the prompt method in use.

\subsubsection{Adaptive token merging (AToM)}

We explain how we bring the above insights into practice through two compute-skipping techniques. We first consider the data-efficient compute-skipping method via token merging. Conventional token merging (ToMe)~\cite{token_merging} reduces the number of tokens by \emph{uniformly} merging \( n \) redundant tokens per layer, controlled by a fixed scheduler \( r \). The scheduler function \( r(l)
\rightarrow n \) is applied to each layer \( l \), and according to \cite{token_merging}, it merges all tokens, including the prompts added by the token selection process. However, insights from~\cref{fig:vitanalysis} and our additional analysis highlight two major problems.

\begin{wrapfigure}{r}{0.48\linewidth}
   \vspace{-5mm}
   \includegraphics[width=\linewidth]{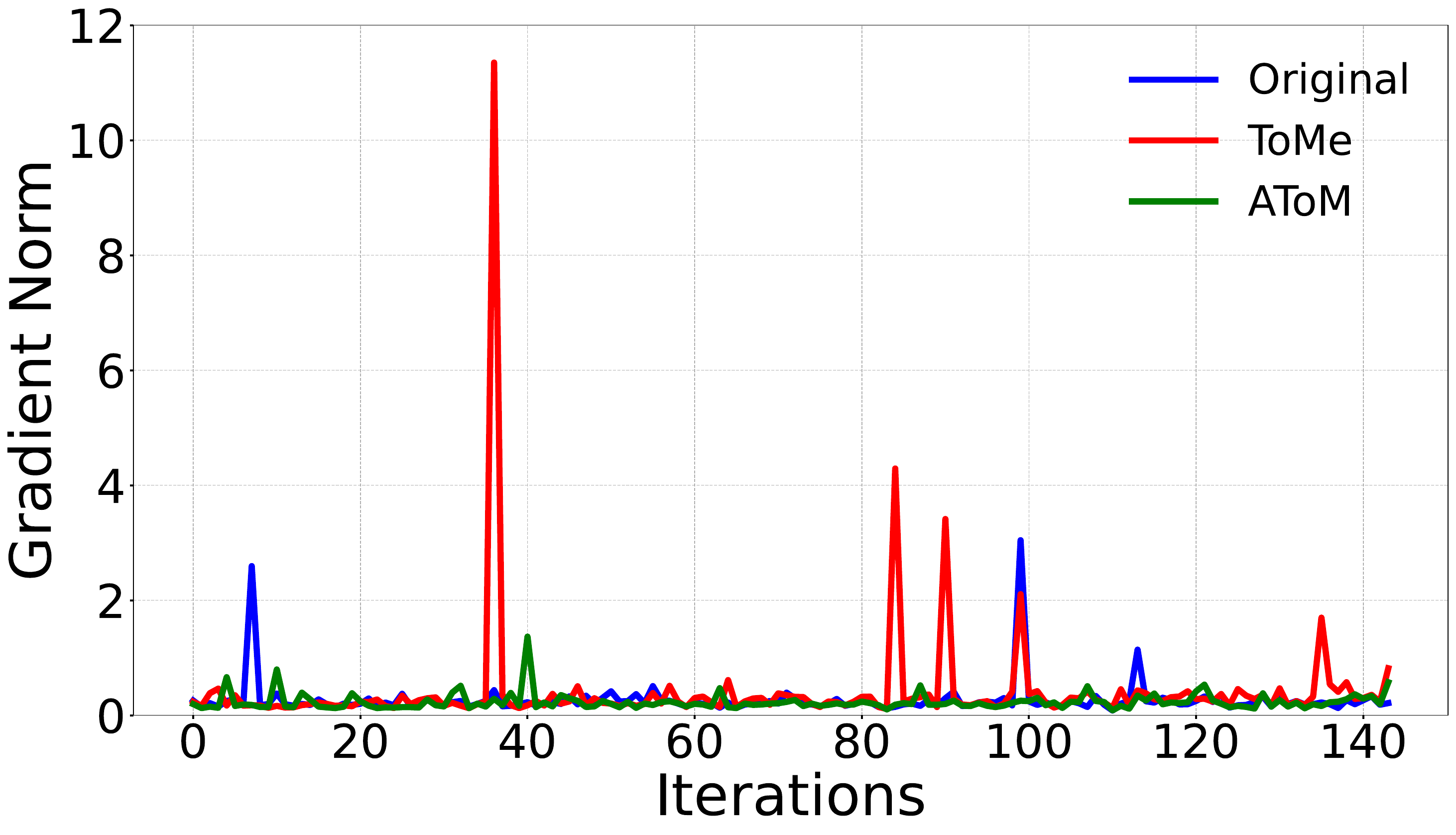}
   \vspace{-5mm}
   \caption{The norm of gradient with respect to the prompt during training Split ImageNet-R (10 tasks) when AToM and ToMe (Conventional token merging) are applied to L2P with ViT-L.}
   \label{fig:promptgrad}
\end{wrapfigure}

First, there is a \emph{loss of task-specific information}. The prompt tokens in CL carry essential task-related information. However, ToMe indiscriminately combines these prompt tokens with non-prompt tokens,
diminishing their intrinsic value. According to our empirical data in~\cref{fig:promptgrad}, this approach can cause gradient explosions in the prompt tokens even with gradient clipping, leading to learning instability.

Second, there is a \emph{lack of layer-specific adaptability}. ToMe does not account for the disparity between shallow and deep layers, treating the importance of all layers uniformly. Therefore, there is a risk of excessive loss of valuable information in shallow layers, which are mainly responsible for adaptability to diverse sequential tasks.

Our \emph{adaptive token merging} (\atom; Algorithm~\ref{alg:atom}) addresses the loss of task-specific information by excluding prompt tokens during token merging, thereby maintaining their specificity. To enhance the effect of prompts and mitigate catastrophic forgetting, \atom uses a new progressive scheduler \( r'(l) \rightarrow n' \), which dynamically adjusts the number of tokens to merge based on layer depth, as follows:
\begin{align}
r'(l) = \min(\delta \times (l - 1), r_{\max}),
\end{align}
where \( l \) denotes the layer index, \( \delta \) is the step change in the
number of tokens to merge, defined as \( \frac{r_{\max}}{\textit{L} - 1} \)
(with \( L \) being the number of layers), and \( r_{\max} \) is the maximum
number of tokens to merge (by default, \( 2 \times n \)). With this
\(r'(l)\), merging occurs more aggressively in deeper layers than in shallow
ones, preserving important task-related representations.

\subsubsection{Adaptive layer dropping (ALD)}

Inspired by insights from~\cref{fig:vitanalysis} and prior work on progressive layer dropping (PLD)~\cite{progressive_drop} in the NLP domain, we propose \emph{adaptive layer dropping} (\ald;~\Cref{alg:ald}) with two key features: (1) the dropping schedule considers both temporal and spatial dimensions, and (2) it manages to drop layers non-uniformly to preserve critical representation features in shallower layers.\footnote{The spatial dimension refers to model layers, each processing input features at different levels of abstraction.} On the contrary, PLD considers only the temporal aspect and does not differentiate between layers when dropping. This
results in poorer performance compared to \ald, as shown in~\Cref{tab:algorithm_validation}.

\ald prioritizes retaining shallow layers that contain richer information essential for model performance, especially after token merging. Thus, instead of operating on its schedule parameters, \ald leverages feedback from \atom, specifically the count of merged tokens at each layer, to guide layer-dropping decisions. The layer-keeping probability \(\theta_{t,l}\) is defined as:
\begin{align}
\theta_{t,l} = \left(\alpha(l)\times( (1 - \bar{\theta})\exp(-\gamma \cdot t) + \bar{\theta})\right),
\end{align}
where \(\theta_{t,l}\) is the probability of keeping layer \(l\) at time step
\(t\), \(\bar{\theta}\) is the minimum probability, \(\gamma\) controls the
decay rate, and \(\alpha(l)\) is the adjustment factor for layer \(l\), defined as:
\begin{align}
\alpha(l) = 
\begin{cases}
\alpha & \text{if } (n(l) - n'(l)) \geq \tau \\
1 & \text{if } (n(l) - n'(l)) < \tau.
\end{cases}
\end{align}

\(\alpha(l)\) quantifies the degree of token merging performed by \atom~in layer \(l\). \(n(l)\) represents the original number of tokens, and \(n'(l)\) denotes the number of tokens remaining after merging. When the number of merged tokens surpasses the threshold \(\tau\), \ald adjusts the layer-dropping probability based on \(\alpha\). Since deeper layers tend to merge more tokens with \atom, \ald is more likely to exceed \(\tau\) in deeper layers and drop more aggressively. These parameters should be tuned to balance between efficiency and the preservation of nuanced information contained in the merged tokens. We set \(\alpha\) to 0.9, with \(\tau\) set to 16 for ViT-L, 12 for ViT-B, and 8 for ViT-Ti, respectively.

\begin{wrapfigure}{r}{0.5\linewidth}
   \centering
   \vspace{-6.6mm}
   \includegraphics[width=\linewidth]{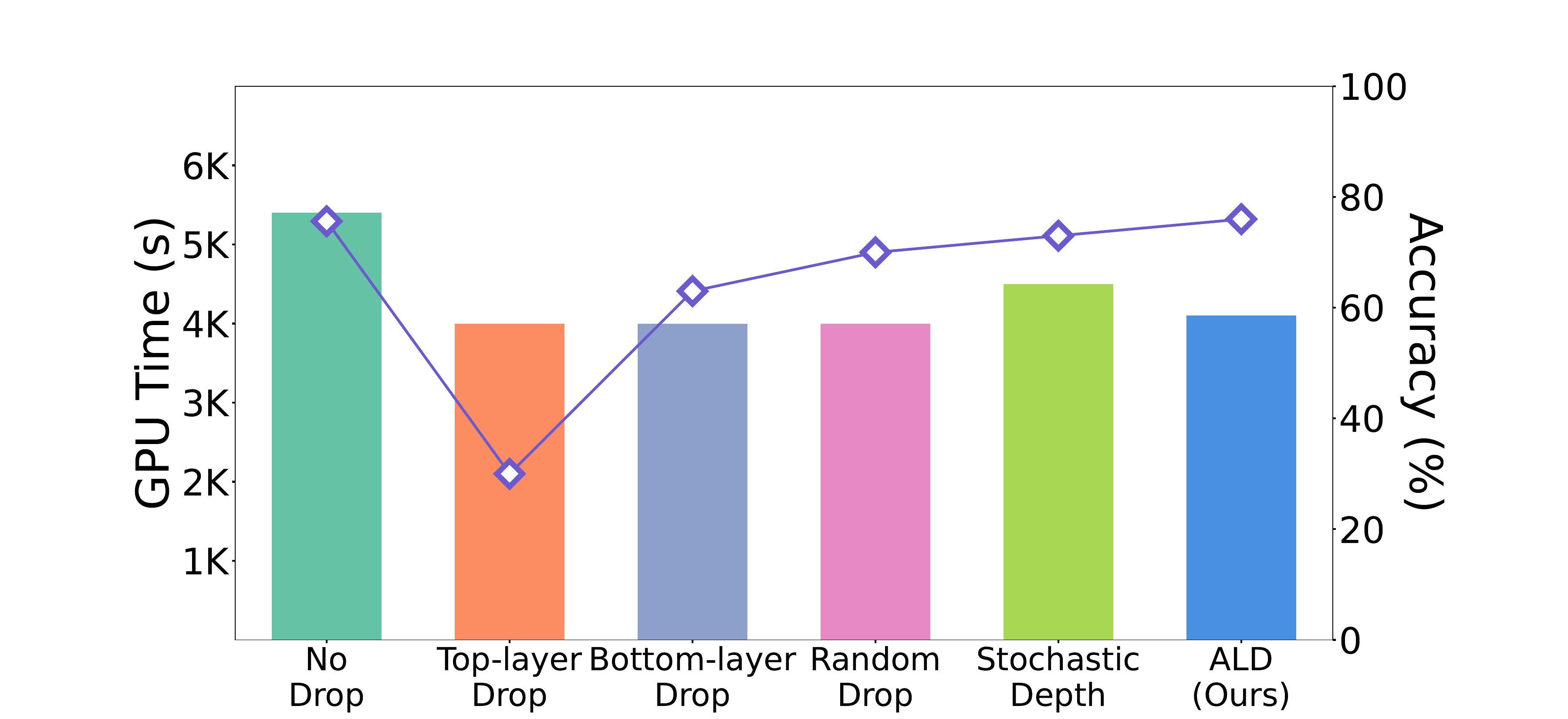}
   \vspace{-6.6mm}
   \caption{Comparing various layer-dropping strategies using L2P with the ViT-L backbone on Split ImageNet-R (10 tasks). The bar and marker are GPU time and final average accuracy, respectively.}
   \label{fig:layerdrop}
\end{wrapfigure}

\textbf{Comparison with various layer-dropping strategies.\;}
We compare \ald against multiple layer-dropping strategies that skip computation across different portions of backbone layers. We report the training wall-clock GPU time and final average accuracy in~\cref{fig:layerdrop}, using L2P with the ViT-L backbone on Split ImageNet-R (10 tasks). Specifically, Top-layer Drop and Bottom-layer Drop statically drop the first and last 25\% of layers, respectively. Random Drop randomly skips 25\% of layers across the network, while Stochastic Depth applies a linear decay schedule over Random Drop. For reference, we include No Drop, which performs no layer dropping and yields 75\% accuracy. To account for the stochastic nature of compared layer dropping methods, such as Stochastic Depth, we measure end-to-end wall-clock GPU time. Top-layer Drop performs the worst, achieving only 30\% accuracy. Bottom-layer Drop (63\%), Random Drop (70\%), and Stochastic Depth (72\%) also face noticeable accuracy degradation despite reduced GPU time. 

In contrast, \ald maintains 75\% accuracy---on par with No Drop---while achieving GPU time savings comparable to Random Drop. This highlights that \ald's non-uniform strategy accounts for each layer's essential spatial representations more effectively. \ours triggers \atom and \ald for each task insertion. Much like our prompt-selection optimization, \atom consistently enhances time and memory efficiency. In contrast, \ald solely
contributes to reducing time costs since it operates with the layer-keeping probability \(\theta_{t,l}\) initialized at 1.0, \ie, no layer dropping.

\subsubsection{Discussion}

\parlabel{Alternative adaptation approaches.\;}
Prompt-based adaptation represents an effective paradigm for continual learning, particularly in resource-constrained settings, owing to its modularity and small parameter footprint. Unlike full fine-tuning, prompting keeps the backbone entirely frozen while allocating a small number of learnable parameters per task. This not only preserves prior knowledge and prevents catastrophic forgetting but also drastically reduces memory usage, which is a critical factor for edge devices. In comparison to alternatives such as LoRA, RanPAC, and adapter tuning, prompting exhibits a superior balance between efficiency and performance, as demonstrated by our integration of \ours into multiple state-of-the-art prompt-based CL methods. 

\parlabel{Compact backbone networks (\eg, CNNs).\;} Although CNNs are generally more compact than vision transformers (ViTs), they consistently underperform in the CL setting, particularly under tight resource constraints. Our preliminary studies in~\Cref{s:preliminary_empirical_study} demonstrate that state-of-the-art CNN-based CL methods, such as BudgetCL~\cite{budgetcl} and MEMO~\cite{memo}, achieve significantly lower accuracy compared to ViT-based approaches.

\section{Experiments}
\label{sec:experiments}

We focus on the popular disjoint-CIL setup, where each task comprises a distinct set of non-overlapping classes and samples from old classes are not given in future tasks~\cite{icarl, castro2018eccv, gepperthH16}. In this
section, we demonstrate that (1) \ours enhances resource efficiency in prompt-based rehearsal-free CL across various backbone model sizes and (2) its core components, \ald and \atom, are applicable even when CL methods do
not employ prompting.

\parlabel{Data generation.\;}
To organize task streams, we use three image classification datasets: CIFAR-100 (100 classes)~\cite{cifar}, ImageNet-R (200 classes)~\cite{imagenet_r}, and PlantDisease (38 classes)~\cite{plantdisease}. Out of these datasets, ImageNet-R is known for exhibiting much higher intra-class variability and an uneven class-size distribution among images. We divide CIFAR-100 and ImageNet-R into 10 tasks to create \textbf{Split CIFAR-100} (\ie, 10 classes per task) and \textbf{Split ImageNet-R} (\ie, 20 classes per task), respectively~\cite{l2p, dualp,codap}. For PlantDisease, we drop 3 plant disease classes with very few images and organize the remaining 35 classes into 7 tasks to create \textbf{Split PlantDisease} (\ie, 5 classes per task). We also present additional results with varying task lengths (\eg, 5 tasks and 20 tasks), and an additional Split CUB-200 dataset~\cite{CUB200} in~\Cref{s:task_scale} and~\Cref{s:newdatasets}, respectively.

\parlabel{Methods.\;}
We employ \textbf{L2P}~\cite{l2p}, \textbf{DualPrompt}~\cite{dualp}, \textbf{CODA-Prompt}~\cite{codap}, \textbf{HiDe-Prompt}~\cite{hidep}, \textbf{ConvPrompt}~\cite{convp}, \textbf{LAE}~\cite{lae}, and \textbf{OVOR}~\cite{ovor} as representative prompting methods. These methods capitalize on ImageNet pre-trained models as backbones: ViT-L (307M), ViT-B (86M), and ViT-Ti (5.8M)---ViT-Ti is \textit{one of the smallest vision transformer models} to our knowledge. Detailed settings for all methods are provided in~\Cref{appendix:details}.

\parlabel{Hardware and metrics.\;}
We use an NVIDIA RTX 3090 to cover a wide range of memory capacities while maintaining consistent computational power. Specifically, we limit the GPU memory to emulate resource-constrained devices, ranging from 1GB (compatible with Raspberry Pi) to 16GB (compatible with NVIDIA Jetson AGX Xavier). For computational efficiency, we measure \textit{iteration time}, which is linearly correlated with GPU energy usage~\cite{miro}. We adopt the \textit{final average accuracy} (higher is better) and \textit{forgetting} (lower is better) from various previous works~\cite{dualp, l2p, codap, hidep}. In~\Cref{appendix:repedgedevices}, we further evaluate \ours directly on NVIDIA Jetson TX2 with 8GB of memory, although it is applicable beyond this specific device.

\subsection{Main results}
\label{subsec:results}

\begin{table*}[t!]
\centering
\caption{
Accuracy and computational cost of all competing methods on three datasets. We report the final average accuracy, iteration time, and memory usage, both without and with \ours. Iteration time and memory usage are presented as absolute values. For the results without \ours, parentheses indicate how many times higher the value is relative to its counterpart with \ours.}
\vskip 0.15in
\setlength{\tabcolsep}{3pt}
\resizebox{\textwidth}{!}{%
\begin{tabular}{l l
                | c c  c c  c c
                | c c
                | c c}
\toprule
& &
\multicolumn{6}{c}{\textbf{Final average accuracy} (\(\uparrow\))} &
\multicolumn{2}{c}{\textbf{Iter. time (ms)}} &
\multicolumn{2}{c}{\textbf{Mem. (GB)}} \\

\textbf{Model} & \textbf{Method} &
\multicolumn{2}{c}{Split CIFAR-100} &
\multicolumn{2}{c}{Split ImageNet-R} &
\multicolumn{2}{c}{Split PlantDisease} \\

\cmidrule(lr){3-4}\cmidrule(lr){5-6}\cmidrule(lr){7-8}\cmidrule(lr){9-10}\cmidrule(lr){11-12}
& &
w/o \ours & w/ \ours &
w/o \ours & w/ \ours &
w/o \ours & w/ \ours &
w/o \ours & w/ \ours &
w/o \ours & w/ \ours \\
\cmidrule(lr){1-2}\cmidrule(lr){3-8}
\cmidrule(lr){9-10}\cmidrule(lr){11-12}
\multirow{8}{*}{ViT-L}
& L2P            & 88.2\(\pm\)0.3 & 87.0\(\pm\)1.3 & 75.6\(\pm\)1.0 & 75.3\(\pm\)1.5 & 75.9\(\pm\)3.1 & 81.1\(\pm\)3.9 & 447(\(\times\)1.9) & 240 & 6.5(\(\times\)1.4) & 4.5 \\
& DualPrompt     & 86.3\(\pm\)0.3 & 85.1\(\pm\)1.3 & 71.2\(\pm\)0.6 & 70.6\(\pm\)0.9 & 75.1\(\pm\)1.1 & 78.2\(\pm\)2.9 & 424(\(\times\)2.0) & 208 & 5.9(\(\times\)1.4) & 4.3 \\
& CODA-Prompt    & 85.4\(\pm\)0.4 & 84.3\(\pm\)1.7 & 74.4\(\pm\)0.6 & 73.5\(\pm\)1.9 & 75.0\(\pm\)2.0 & 74.1\(\pm\)2.7 & 568(\(\times\)1.3) & 441 & 13.2(\(\times\)1.2) & 11.0 \\
& HiDe-Prompt    & 93.4\(\pm\)0.3 & 92.8\(\pm\)1.1 & 78.7\(\pm\)0.2 & 78.0\(\pm\)1.2 & 98.6\(\pm\)0.4 & 98.3\(\pm\)0.8 & 413(\(\times\)1.8) & 227 & 7.3(\(\times\)1.2) & 6.3 \\
& ConvPrompt     & 89.2\(\pm\)0.5 & 89.0\(\pm\)0.9 & 79.1\(\pm\)0.1 & 78.5\(\pm\)0.8 & 98.4\(\pm\)0.6 & 98.6\(\pm\)0.7 & 560(\(\times\)1.3) & 417 & 5.4(\(\times\)1.3) & 4.1 \\
& LAE-Prefix10  & 84.9\(\pm\)1.3 & 84.5\(\pm\)1.5 & 70.5\(\pm\)1.9 & 70.2\(\pm\)1.7 & 73.3\(\pm\)0.5 & 72.9\(\pm\)0.9 & 185(\(\times\)1.1) & 170 & 5.1(\(\times\)1.1) & 4.8 \\
& OVOR           & 86.2\(\pm\)0.6 & 85.8\(\pm\)0.9 & 75.4\(\pm\)0.7 & 74.3\(\pm\)1.3 & 75.4\(\pm\)0.8 & 74.9\(\pm\)1.0 & 467(\(\times\)1.3) & 362 & 7.5(\(\times\)1.1) & 6.8 \\
& UpperBound     & 94.1\(\pm\)0.1 & --            & 85.2\(\pm\)0.2 & --            & 99.6\(\pm\)0.3 & --            & 427              & --  & 9.8              & -- \\
\midrule
\multirow{8}{*}{ViT-B}
& L2P            & 84.5\(\pm\)0.8 & 83.4\(\pm\)1.8 & 59.4\(\pm\)0.8 & 58.7\(\pm\)1.2 & 64.4\(\pm\)2.6 & 64.0\(\pm\)2.4 & 143(\(\times\)1.4) & 102 & 2.3(\(\times\)1.2) & 2.0 \\
& DualPrompt     & 85.1\(\pm\)0.2 & 84.9\(\pm\)0.9 & 68.1\(\pm\)0.1 & 67.0\(\pm\)1.0 & 78.0\(\pm\)1.5 & 77.2\(\pm\)1.9 & 133(\(\times\)1.3) & 101 & 2.1(\(\times\)1.2) & 1.8 \\
& CODA-Prompt    & 83.3\(\pm\)0.9 & 82.1\(\pm\)1.9 & 59.0\(\pm\)0.3 & 58.3\(\pm\)1.8 & 71.4\(\pm\)0.8 & 70.9\(\pm\)2.7 & 164(\(\times\)1.1) & 151 & 6.9(\(\times\)1.2) & 5.7 \\
& HiDe-Prompt    & 88.5\(\pm\)0.7 & 89.2\(\pm\)1.6 & 64.5\(\pm\)0.5 & 64.4\(\pm\)1.0 & 94.9\(\pm\)0.4 & 94.5\(\pm\)0.8 & 130(\(\times\)1.7) & 76  & 4.1(\(\times\)1.7) & 2.4 \\
& ConvPrompt     & 87.6\(\pm\)0.5 & 87.0\(\pm\)1.4 & 67.6\(\pm\)0.6 & 67.1\(\pm\)1.2 & 94.3\(\pm\)0.8 & 94.5\(\pm\)0.9 & 381(\(\times\)1.4) & 279 & 2.2(\(\times\)1.1) & 2.0 \\
& LAE-Prefix10  & 83.5\(\pm\)0.8 & 84.2\(\pm\)1.3 & 61.3\(\pm\)0.6 & 61.0\(\pm\)0.9 & 70.8\(\pm\)0.6 & 70.4\(\pm\)1.1 &  57(\(\times\)1.1) & 50  & 1.9(\(\times\)1.1) & 1.8 \\
& OVOR           & 84.2\(\pm\)0.7 & 83.9\(\pm\)1.0 & 61.0\(\pm\)0.6 & 60.6\(\pm\)1.0 & 67.2\(\pm\)0.5 & 66.9\(\pm\)0.9 & 134(\(\times\)1.1) & 123 & 3.1(\(\times\)1.1) & 3.0 \\
& UpperBound     & 92.0\(\pm\)0.2 & --            & 81.4\(\pm\)0.1 & --            & 99.6\(\pm\)0.4 & --            & 143              & --  & 3.2              & -- \\
\midrule
\multirow{8}{*}{ViT-Ti}
& L2P            & 60.3\(\pm\)0.2 & 59.3\(\pm\)1.7 & 41.2\(\pm\)0.7 & 40.3\(\pm\)1.0 & 56.1\(\pm\)2.9 & 55.2\(\pm\)2.9 & 34(\(\times\)1.1) & 30 & 0.5(\(\times\)1.1) & 0.5 \\
& DualPrompt     & 62.9\(\pm\)0.5 & 62.5\(\pm\)0.9 & 43.6\(\pm\)0.8 & 42.7\(\pm\)1.0 & 64.0\(\pm\)1.0 & 62.9\(\pm\)1.9 & 34(\(\times\)1.1) & 31 & 0.4(\(\times\)1.1) & 0.4 \\
& CODA-Prompt    & 67.0\(\pm\)0.4 & 65.9\(\pm\)1.8 & 49.3\(\pm\)0.8 & 48.2\(\pm\)2.0 & 66.8\(\pm\)0.8 & 65.5\(\pm\)2.8 & 36(\(\times\)1.2) & 30 & 2.8(\(\times\)1.1) & 2.8 \\
& HiDe-Prompt    & 72.2\(\pm\)0.5 & 72.5\(\pm\)1.0 & 44.8\(\pm\)0.8 & 45.2\(\pm\)1.4 & 92.6\(\pm\)0.7 & 93.3\(\pm\)1.5 & 23(\(\times\)1.2) & 20 & 0.5(\(\times\)1.1) & 0.4 \\
& ConvPrompt     & 69.1\(\pm\)0.8 & 68.5\(\pm\)1.7 & 49.8\(\pm\)0.8 & 49.1\(\pm\)1.9 & 92.4\(\pm\)0.6 & 93.4\(\pm\)0.9 & 85(\(\times\)1.9) & 45 & 0.5(\(\times\)1.1) & 0.5 \\
& LAE-Prefix10  & 61.3\(\pm\)0.6 & 61.3\(\pm\)1.4 & 49.7\(\pm\)0.8 & 49.6\(\pm\)1.3 & 66.8\(\pm\)0.7 & 66.0\(\pm\)1.5 & 15(\(\times\)1.3) & 12 & 0.4(\(\times\)1.1) & 0.4 \\
& OVOR           & 67.7\(\pm\)0.8 & 67.0\(\pm\)1.3 & 49.8\(\pm\)0.7 & 49.3\(\pm\)1.2 & 70.3\(\pm\)0.6 & 71.6\(\pm\)1.7 & 35(\(\times\)1.1) & 32 & 1.0(\(\times\)1.1) & 0.9 \\
& UpperBound     & 83.0\(\pm\)0.1 & --            & 77.4\(\pm\)0.2 & --            & 98.0\(\pm\)0.9 & --            & 42               & -- & 0.6               & -- \\
\bottomrule
\end{tabular}}%
\label{tab:comparison}
\vskip -0.1in
\end{table*}

\Cref{tab:comparison} presents the efficacy of \ours when integrated into prompt-based CL methods across various ViT backbones (ViT-L, ViT-B, and ViT-Ti) and datasets. For a direct comparison of these methods under their original setups, please refer to~\Cref{s:sotaoriginal}.

\noindent\textbf{Resource efficiency.\;}
Integrating \ours into existing prompt-based CL methods substantially reduces computational overhead and memory usage across all model configurations. As shown in~\Cref{tab:comparison}, baseline methods without \ours consume up to \(1.4\times\), \(1.7\times\), and \(1.1\times\) more GPU memory than those with \ours for ViT-L, ViT-B, and ViT-Ti, respectively. Although the ViT backbones remain frozen, these models still perform backward passes to optimize prompts and classification heads, requiring memory to store large intermediate activations. \ours mitigates these memory demands by selectively merging tokens (\atom) and adaptively dropping layers (\ald), thus reducing the memory footprint of intermediate data.

In addition, baseline methods without \ours use up to \(2.0\times\), \(1.7\times\), and \(1.9\times\) more training time than those with \ours for ViT-L, ViT-B, and ViT-Ti, respectively. These improvements in computational efficiency directly translate into proportionally lower energy consumption. The gains are more marked in ViT-L, as its larger size offers greater room for optimizing computation cost.

\noindent\textbf{Accuracy.\;}
While all baseline methods were originally designed and evaluated on a ViT-B backbone, our study broadens their applicability by exploring cost-accuracy trade-offs in on-device CL across a diverse range of ViT architectures. When augmented with our techniques, these methods achieve notable gains in resource efficiency with only marginal accuracy degradation: 0.0--1.2\% on Split CIFAR-100, 0.1--1.1\% on Split ImageNet-R, and 0.0--0.8\% on Split PlantDisease. In some cases, accuracy even improves with \ours; for example, the accuracy of L2P with ViT-L on Split PlantDisease increases from 75.9\% to 81.1\% when combined with \ours. For completeness, the forgetting metric is reported in~\Cref{s:fullcomparison}.

\subsection{Ablation studies}
\label{subsec:ablation}

\begin{wraptable}{r}{0.48\linewidth}
\vspace{-6.6mm}
\caption{Ablation study of \ours's components, demonstrating their contributions to both resource efficiency and accuracy.}
\centering
\setlength{\tabcolsep}{2pt}
\vskip 0.15in
\resizebox{\linewidth}{!}{
\begin{tabular}{lccccc}
\toprule
Ablated components & Acc. (\(\uparrow\)) & Fgt. (\(\downarrow\))& Iter. time (ms) & Mem. (GB) \\
\midrule
\ours-L2P & 75.3\(\pm\)1.5 & 3.6\(\pm\)0.3 & 240 & 4.5\\
\midrule
w/o (\atom + \ald)
&74.8\(\pm\)0.8 & 4.0\(\pm\)0.8 & 349 & 5.5\\
w/o (\(f_{\text{efficient}}\) + \ald)
&74.9\(\pm\)0.8 & 3.5\(\pm\)0.8 & 419 & 5.6\\
w/o (\(f_{\text{efficient}}\) + \atom)
&74.5\(\pm\)0.8 & 3.6\(\pm\)0.7 & 401 & 6.5\\
w/o \ald
& 74.2\(\pm\)0.8 & 4.0\(\pm\)0.8   & 270 & 4.5 \\
w/o \atom
&74.5\(\pm\)0.4 & 3.4\(\pm\)0.5 & 303 & 5.5\\
w/o \(f_{\text{efficient}}\)
& 74.6\(\pm\)0.4 & 2.6\(\pm\)0.5 & 326 & 4.8 \\
\bottomrule
\end{tabular}
}
\vspace{-5mm}
\label{tab:component_ablation}
\end{wraptable}

We validate our proposed techniques and the chosen hyperparameters, primarily using \ours integrated into L2P with the ViT-L backbone (denoted as REP-L2P) on Split ImageNet-R (10 tasks). See~\Cref{s:extend_ablation} for additional results.

\parlabel{Component ablation.\;}
In~\Cref{tab:component_ablation}, we ablate the components of \ours to assess their individual contributions. Each row reports performance when the corresponding component is removed. All components contribute substantially to reducing computation time and memory usage, with \ald affecting only computation time, as expected.

\begin{wraptable}{r}{0.48\linewidth}
\vspace{-6.6mm}
\caption{Effect of using (1) conventional acceleration methods, (2) intermediate layer-based prompt selection method, and (3) random drop with diverse drop ratios, compared to our proposed methods.}
\centering
\setlength{\tabcolsep}{2pt}
\vskip 0.15in
\resizebox{\linewidth}{!}{
\begin{tabular}{lcccc}
\toprule
Method & Acc. (\(\uparrow\)) & Fgt. (\(\downarrow\)) & Iter. time (ms) & Mem. (GB) \\
\midrule
\multicolumn{5}{l}{\textit{(1) Algorithm validation}} \\
\cmidrule(lr{5pt}){1-5}
\ours-L2P & 75.3\(\pm\)1.5 & 3.6\(\pm\)0.3 & 240 & 4.5 \\
w/ ToMe & 70.2\(\pm\)0.7 & 2.6\(\pm\)0.9 & 275 & 3.7 \\
w/ PLD & 73.3\(\pm\)0.7 & 3.9\(\pm\)0.7 & 259 & 4.5 \\
\cmidrule(lr{5pt}){1-5}
\multicolumn{5}{l}{\textit{(2) Surrogate prompt selection validation}} \\
\cmidrule(lr{5pt}){1-5}
Surrogate & 75.3\(\pm\)1.5 & 3.6\(\pm\)0.3 & 240 & 4.5 \\
Interm. (4) & 74.0\(\pm\)0.8 & 3.9\(\pm\)0.9 & 264 & 4.3 \\
Interm. (8) & 74.3\(\pm\)0.7 & 3.5\(\pm\)0.9 & 285 & 4.3 \\
Interm. (12) & 74.6\(\pm\)0.7 & 2.6\(\pm\)0.8 & 307 & 4.3 \\
\cmidrule(lr{5pt}){1-5}
\multicolumn{5}{l}{\textit{(3) Comparison with FIM-based token merging}} \\
\cmidrule(lr{5pt}){1-5}
\atom & 74.5\(\pm\)0.8 & 3.5\(\pm\)0.8 & 419 & 5.6 \\
FIM-based & 74.7\(\pm\)0.9 & 3.8\(\pm\)0.4 & 264 & 6.2 \\
\cmidrule(lr{5pt}){1-5}
\multicolumn{5}{l}{\textit{(4) Comparison with random drop}} \\
\cmidrule(lr{5pt}){1-5}
\ald & 75.8\(\pm\)0.5 & 3.5\(\pm\)0.7 & 401 & 6.5 \\
Random-20\% & 71.7\(\pm\)0.4 & 5.8\(\pm\)0.8 & 409 & 6.5 \\
Random-25\% & 70.6\(\pm\)0.4 & 6.4\(\pm\)1.1 & 398 & 6.5 \\
Random-30\% & 68.8\(\pm\)0.6 & 7.1\(\pm\)1.5 & 385 & 6.5 \\
Random-50\% & 65.4\(\pm\)0.8 & 9.4\(\pm\)1.8 & 363 & 6.5 \\
\bottomrule
\end{tabular}
}
\vspace{-5mm}
\label{tab:algorithm_validation}
\end{wraptable}

\parlabel{Algorithm validation.\;}
Conventional token merging (ToMe)~\cite{token_merging} and layer dropping (PLD)~\cite{progressive_drop} are specifically designed to accelerate the training of transformer-based models across various domains.
To validate the importance of incorporating our adaptive techniques instead in CL, we first evaluate how \ours performs in case \atom and \ald are replaced with ToMe and PLD, respectively. 
The results are presented in (1) of~\Cref{tab:algorithm_validation}. Although applying ToMe or PLD improves resource efficiency over the native L2P baseline in~\Cref{tab:comparison}, it results in an accuracy drop of 5.1\% or 2.0\%, respectively, compared to using our techniques.

\parlabel{Surrogate prompt selection validation.\;}
We further validate the effectiveness of our lightweight surrogate-based prompt selection method by comparing it with an alternative that uses the CLS token from intermediate layers of ViT-L (layers 4, 8, and 12). As shown in (2) of~\Cref{tab:algorithm_validation}, while using intermediate layers eliminates the need for a separate surrogate, it substantially increases training time, since even a partial ViT-L forward pass is more computationally demanding than running the full surrogate model. Accuracy also drops, with shallower-layer cases showing greater degradation, likely due to their limited access to global context.

\parlabel{Comparison with FIM-based token merging.\;}
We also investigate whether additional gradient information, \eg through the Fisher Information Matrix (FIM)~\cite{ewc}, could serve as an alternative to \atom. Specifically, we compare \atom with a FIM-based, gradient-informed merging strategy using L2P with ViT-L on Split ImageNet-R (10 tasks), as shown in (3) of~\Cref{tab:algorithm_validation}. Although this approach yields modest accuracy gains (0.1--0.2\%), it incurs roughly a 10\% increase in training time and memory usage. By contrast, \atom preserves prompt gradients without such overhead, offering a more balanced trade-off between efficiency and forgetting mitigation. This result does not point to a fundamental limitation of FIM-based merging, but rather underscores the difficulty of achieving practical efficiency improvements in its current form.

\parlabel{Comparison with random drop across varying aggressiveness levels.\;}
To evaluate \ald against Random Drop with varying aggressiveness, we applied dropping rates of 20\%, 25\%, 30\%, and 50\% using L2P with ViT-L on on Split ImageNet-R (10 tasks). As shown in (4) of~\Cref{tab:algorithm_validation}, higher drop rates reduce training time below that of ALD but lead to substantial accuracy degradation (65--71\%), making them impractical for continual learning. In contrast, \ald dynamically adjusts the drop ratio, maintaining resource efficiency while minimizing accuracy loss.

\begin{wraptable}{r}{0.48\linewidth}
\vspace{-6.8mm}
\centering
\caption{REP over varying \# of merged tokens (\(n\)) and \% of keep ratio (\(\theta\)).}
\setlength{\tabcolsep}{2pt}
\vskip 0.15in
\resizebox{\linewidth}{!}{
    \begin{tabular}{ccccc}
        \toprule
        $n$  (w/ \(\theta\)=0.5) & Acc. (\(\uparrow\)) & Fgt. (\(\downarrow\)) & Iter. time (ms) & Mem. (GB)\\
        \midrule
        1 & 75.5\(\pm\)0.7 & 3.5\(\pm\)0.7 & 268 & 5.5\\
        2 & 75.2\(\pm\)0.7 & 3.7\(\pm\)0.7 & 265 & 5.5\\
        4 & 75.3\(\pm\)0.3 & 3.3\(\pm\)0.5 & 256 & 5.2\\
        6 & 75.1\(\pm\)0.6 & 3.8\(\pm\)0.8 & 253 & 4.8\\
        8 & 75.3\(\pm\)1.5 & 3.6\(\pm\)0.3 & 240 & 4.5\\
        10 & 73.6\(\pm\)1.7 & 4.7\(\pm\)1.1 & 228 & 4.1\\
        \midrule
        \(\theta\)  (w/ $n$=8) & Acc. (\(\uparrow\)) & Fgt. (\(\downarrow\)) & Iter. time (ms) & Mem. (GB)\\
        \midrule       
        0.1 & 72.9\(\pm\)1.4 & 4.4\(\pm\)1.5 & 217 & 4.5\\
        0.2 & 73.2\(\pm\)1.2 & 4.4\(\pm\)1.2 & 223 & 4.5\\
        0.3 & 74.4\(\pm\)1.2 & 4.1\(\pm\)0.8 & 228 & 4.5\\
        0.4 & 74.7\(\pm\)1.0 & 3.9\(\pm\)0.5 & 235 & 4.5\\
        0.5 & 75.3\(\pm\)1.5 & 3.6\(\pm\)0.3 & 240 & 4.5\\
        0.6 & 74.8\(\pm\)0.9 & 3.6\(\pm\)0.9 & 255 & 4.5\\
        0.7 & 74.4\(\pm\)0.6 & 3.5\(\pm\)0.5 & 270 & 4.5\\
        0.8 & 74.2\(\pm\)0.7 & 3.8\(\pm\)0.9 & 277 & 4.5\\
        0.9 & 74.3\(\pm\)0.5 & 3.6\(\pm\)0.8 & 282 & 4.5\\
        \bottomrule
    \end{tabular}
}
\vspace{-6.6mm}
\label{tab:adjustratio}
\end{wraptable}

\parlabel{AToM and ALD intensity.\;}
\Cref{tab:adjustratio} shows the effects of varying intensities of \atom and \ald, focusing on the number of merged tokens (\(n\)) in \atom and the keep ratio (\(\theta\)) in \ald. \atom appears to maintain stable accuracy trends even as more tokens are merged. In contrast, in the case of \ald, a lower keep ratio appears to improve resource efficiency by reducing training time. However, it can markedly impair model accuracy if the ratio is too low. Overall, when used with carefully selected hyperparameters, \atom and \ald can effectively balance resource efficiency and accuracy. Notably, forgetting remains stable when using the default settings (\(n\) = 8 and \(\theta\) = 0.5). In~\Cref{s:extend_ablation}, we extend the above study to two additional prompt-based CL methods, including HiDe-Prompt~\cite{hidep} and ConvPrompt~\cite{convp}. \\

\subsection{Additional studies}
We further assess the efficacy of \ours by applying it to CL methods not covered by~\cref{tab:comparison}, including two additional non-prompting rehearsal-free methods: such as SLCA~\cite{slca} and RanPAC~\cite{ranpac}. Additionally, we compare \ours-L2P with two adapter-based methods, including Online-LoRA~\cite{onlinelora} and ADAM~\cite{adam}. More discussions are provided in~\cref{s:extend_additional} and~\cref{s:additional_related}, respectively.

\subsubsection{Applying \ours to non-prompting rehearsal-free methods}

SLCA fine-tunes the entire network using distinct learning rates for the representation and fully-connected layers, whereas RanPAC employs a non-trainable random projection to the activations from the pre-trained frozen backbone. For SLCA, we retain the original hyperparameters (\eg, the number of epochs per task), and integrate ConvPrompt to enable parameter-efficient fine-tuning (PEFT), as SLCA lacks native PEFT support.
We measure iteration time and memory usage, with ViT-L on Split ImageNet-R (10 tasks).

As shown in~\cref{tab:slca}, SLCA incurs high memory overhead even with ConvPrompt, limiting its compatibility with 8GB edge devices. Applying \ours reduces both training time and memory usage by up to 48\%, making it operational on such resource-constrained devices. Similarly, applying \ours to RanPAC reduces iteration time by up to 37\% (as shown in \Cref{tab:ranpac}) but does not lower memory usage, as random projections from upsampling dominate the high memory cost.

\begin{table}[ht]
\begin{minipage}[t]{0.48\linewidth}
\centering
\caption{\ours on SLCA with ConvPrompt using ViT-L backbone on Split ImageNet-R.}
\scriptsize
\setlength{\tabcolsep}{2pt}
\resizebox{\linewidth}{!}{
\begin{tabular}{lccc}
\toprule
Method &  Acc. (\(\uparrow\)) & Iter. time (ms) & Mem. (GB) \\
\midrule
SLCA & 77.2\(\pm\)1.1 & 1,311 & 9.3 \\
\ours-SLCA & 78.0\(\pm\)1.3 & 686 & 4.8 \\
\bottomrule
\end{tabular}
}
\label{tab:slca}
\end{minipage}
\hfill
\begin{minipage}[t]{0.48\linewidth}
\caption{\ours on RanPAC using ViT-L backbone on Split ImageNet-R.}
\centering
\scriptsize
\setlength{\tabcolsep}{2pt}
\resizebox{\linewidth}{!}{
\begin{tabular}{lcccc}
\toprule
Method &  Acc. (\(\uparrow\)) & Iter. time (ms) & Mem. (GB) \\
\midrule
RanPAC & 82.4\(\pm\)0.6 & 332 & 8.2 \\
\ours-RanPAC & 82.0\(\pm\)1.7 & 210 & 8.2 \\
\bottomrule
\end{tabular}
}
\label{tab:ranpac}
\end{minipage}
\end{table}

\subsubsection{Comparison with adapter-based methods}
\leavevmode

\begin{wraptable}{r}{0.48\linewidth}
\vspace{-6.6mm}
\centering
\caption{Comparison of \ours with Online-LoRA using ViT-L backbone on Split ImageNet-R.}
\setlength{\tabcolsep}{2pt}
\vskip 0.15in
\resizebox{\linewidth}{!}{
        \begin{tabular}{lcccc}
            \toprule
             Method & Acc. (\(\uparrow\)) & Fgt. (\(\downarrow\)) & Iter. time (ms) & Mem. (GB) \\
            \midrule
            \ours-L2P & 75.3\(\pm\)1.5 & 2.8\(\pm\)0.2 & 240 & 4.5 \\
            Online-LoRA & 52.6\(\pm\)1.8 & 21.2\(\pm\)2.6 & 1,739 & 20.9 \\
            \bottomrule
            \end{tabular}}
\label{tab:loracomparison}
\vspace{-6.6mm}
\end{wraptable}

\parlabel{LoRA-based method.\;}
We analyze Online-LoRA using the ViT-L backbone on Split ImageNet-R (10 tasks). As shown in~\cref{tab:loracomparison}, this method introduces substantial memory overhead (>20GB), making it incompatible with the memory specifications of most edge devices. This overhead arises from using replay buffers and multiple forward/backward passes, which significantly inflate both training time and memory usage.

\begin{wraptable}{r}{0.48\linewidth}
\vspace{-6.6mm}
\centering
\caption{Comparison of \ours with ADAM variants using ViT-L backbone on Split ImageNet-R.}
\setlength{\tabcolsep}{2pt}
\vskip 0.15in
\resizebox{\linewidth}{!}{
    \begin{tabular}{lccc}
        \toprule
        Method & Acc. (\(\uparrow\)) & Iter. time (ms) & Mem. (GB) \\
        \midrule
        \ours-L2P & 72.6\(\pm\)0.6 & 240 & 4.50 \\
        \midrule
        VPT-Deep & 66.7\(\pm\)0.4 & 136 & 12.6 \\
        VPT-Shallow & 64.5\(\pm\)0.5 & 125 & 6.9 \\
        SSF & 72.2\(\pm\)0.2 & 147 & 22.1 \\
        Adapter & 62.1\(\pm\)0.1 & 62 & 12.0 \\
        \bottomrule
    \end{tabular}
}
\vspace{-12mm}
\label{tab:ours_adam}
\end{wraptable}

\parlabel{Other adapter-based methods.\;}
Next, we look into ADAM, a representative adapter-based CL method. \Cref{tab:ours_adam} presents results using the ViT-L backbone on Split ImageNet-R (20 tasks). While ADAM variants can reduce training costs through lightweight adapters, they often suffer from large memory footprints that exceed the device limits (\eg, 8GB) or from significantly lower accuracy compared to \ours-L2P.\\

\section{Conclusions}
\label{sec:conclusion}

In this work, we introduce Resource-Efficient Prompting (REP), a framework designed to enhance the computational and memory efficiency of prompt-based rehearsal-free continual learning methods. By incorporating surrogate prompt selection, adaptive token merging and adaptive layer dropping, REP selectively streamlines the learning process while preserving task-specific features, significantly reducing resource consumption without compromising accuracy. Experiments on multiple image classification datasets demonstrate that REP achieves substantial efficiency gains over state-of-the-art ViT-based rehearsal-free methods, making it a practical solution for on-device continual learning.

We focus on a standard continual learning setting where task arrivals are predefined as part of the problem formulation. In this setting, \ours is designed to operate under the given task order. Curriculum learning, which explicitly determines or optimizes the task sequence, addresses a related but distinct problem. Since our study does not include curriculum-based task scheduling, \ours has not been evaluated in scenarios where the task order can be deliberately arranged. Exploring how REP could be combined with curriculum learning strategies is an interesting direction for future research. Moreover, although using a replay buffer may be considered infeasible in real-world scenarios with strict data privacy constraints, on-device CL is already viewed as a privacy-preserving approach, making rehearsal-based \ours a promising direction for future investigation.

One limitation of our work is hyperparameter tuning. Following prior works, we tuned the hyperparameters through grid search with 5-fold cross-validation for each task. While this provides an automatic tuning procedure, it may limit the ``plug-and-play'' applicability of \ours. Empirically, we observed that performance is relatively insensitive to variations in \(\alpha\), making it reasonable to fix \(\alpha\) = 0.9. For \(\tau\), the impact is more dataset- and model-dependent, but we found a consistent positive correlation between backbone size and the effective \(\tau\) value. As future work, we plan to conduct a more detailed study on hyperparameter tuning.

\newpage
\section*{Acknowledgements}
This work was supported by the Institute of Information \& Communications Technology Planning \& Evaluation (IITP) grants (No.~RS-2019-II191906, Artificial Intelligence Graduate School Program (POSTECH); No.~RS-2024-00459797, Development of ML Compiler
Framework for On-device AI; No.~RS-2025-02304554, Efficient and Scalable Framework for AI Heterogeneous Cluster Systems; No.~RS-2022-II220290, Visual Intelligence for Space-Time Understanding and Generation), the National Research Foundation of Korea (NRF) grants (No.~RS-2024-00337559; No.~RS-2024-00354947), and the Electronics and Telecommunications Research Institute (ETRI) grant (No.~25ZS1100, Research on High-Performance Computing to Overcome Limitations of AI), all funded by the Korean government (MSIT).

{
\small
\bibliographystyle{plainnat}
\bibliography{main}
}


\section*{NeurIPS Paper Checklist}

\begin{enumerate}

\item {\bf Claims}
    \item[] Question: Do the main claims made in the abstract and introduction accurately reflect the paper's contributions and scope?
    \item[] Answer: \answerYes{} 
    \item[] Justification: We ensure that the claims made in the abstract and introduction accurately reflect our contributions and scope.

\item {\bf Limitations}
    \item[] Question: Does the paper discuss the limitations of the work performed by the authors?
    \item[] Answer: \answerYes{} 
    \item[] Justification: We discuss the limitations of our method in~\cref{sec:conclusion}, including specific learning environments and learning algorithms that affect the broader applicability of this work. 

\item {\bf Theory assumptions and proofs}
    \item[] Question: For each theoretical result, does the paper provide the full set of assumptions and a complete (and correct) proof?
    \item[] Answer: \answerNA{} 
    \item[] Justification: Our method does not require the full set of assumptions and a complete proof. Our method introduces two scheduling algorithms that are intuitive to understand.

    \item {\bf Experimental result reproducibility}
    \item[] Question: Does the paper fully disclose all the information needed to reproduce the main experimental results of the paper to the extent that it affects the main claims and/or conclusions of the paper (regardless of whether the code and data are provided or not)?
    \item[] Answer: \answerYes{} 
    \item[] Justification: We include detailed explanations of our methods along with pseudo code, how to set hyperparameters, and how to split datasets to create continuous tasks to train in the paper. Nonetheless, we attach the experimental code as a supplementary file for reproducibility.

\item {\bf Open access to data and code}
    \item[] Question: Does the paper provide open access to the data and code, with sufficient instructions to faithfully reproduce the main experimental results, as described in supplemental material?
    \item[] Answer: \answerYes{} 
    \item[] Justification: Reference pre-trained models, learning methods, and datasets are publicly available. Our code is zipped and attached as a supplementary file.

\item {\bf Experimental setting/details}
    \item[] Question: Does the paper specify all the training and test details (e.g., data splits, hyperparameters, how they were chosen, type of optimizer, etc.) necessary to understand the results?
    \item[] Answer: \answerYes{} 
    \item[] Justification: We describe basic experimental methodology in~\cref{sec:experiments} in the main paper, and discuss additional details on implementations and hyperparameters in~\cref{s:setupinfo}.

\item {\bf Experiment statistical significance}
    \item[] Question: Does the paper report error bars suitably and correctly defined or other appropriate information about the statistical significance of the experiments?
    \item[] Answer: \answerYes{} 
    \item[] Justification: We calculated the performance for all experiment 5 times, and reported the mean and standard deviations.

\item {\bf Experiments compute resources}
    \item[] Question: For each experiment, does the paper provide sufficient information on the computer resources (type of compute workers, memory, time of execution) needed to reproduce the experiments?
    \item[] Answer: \answerYes{} 
    \item[] Justification: We discuss relevant details in~\cref{sec:experiments} and~\cref{s:gpupowerusage}.
    
\item {\bf Code of ethics}
    \item[] Question: Does the research conducted in the paper conform, in every respect, with the NeurIPS Code of Ethics \url{https://neurips.cc/public/EthicsGuidelines}?
    \item[] Answer: \answerYes{} 
    \item[] Justification: We comprehensively read the ethics guidelines and confirm that we do not violate anonymity of the review process.

\item {\bf Broader impacts}
    \item[] Question: Does the paper discuss both potential positive societal impacts and negative societal impacts of the work performed?
    \item[] Answer: \answerYes{} 
    \item[] Justification: We discuss about societal impacts in~\cref{s:societal}.
    
\item {\bf Safeguards}
    \item[] Question: Does the paper describe safeguards that have been put in place for responsible release of data or models that have a high risk for misuse (e.g., pretrained language models, image generators, or scraped datasets)?
    \item[] Answer: \answerNA{} 
    \item[] Justification: We do not release any new datasets or models. We provide simple compute-skipping techniques for existing continual learning methods using pre-trainined models.

\item {\bf Licenses for existing assets}
    \item[] Question: Are the creators or original owners of assets (e.g., code, data, models), used in the paper, properly credited and are the license and terms of use explicitly mentioned and properly respected?
    \item[] Answer: \answerYes{} 
    \item[] Justification: We properly credited the creators of the publicly available models and datasets by citing the direct links to the websites or original papers where possible.

\item {\bf New assets}
    \item[] Question: Are new assets introduced in the paper well documented and is the documentation provided alongside the assets?
    \item[] Answer: \answerYes{} 
    \item[] Justification: We include our code as an asset in its original form along with an appropriate level of documentation.

\item {\bf Crowdsourcing and research with human subjects}
    \item[] Question: For crowdsourcing experiments and research with human subjects, does the paper include the full text of instructions given to participants and screenshots, if applicable, as well as details about compensation (if any)? 
    \item[] Answer: \answerNA{} 
    \item[] Justification: We did not utilize crowdsourcing or human subjects.

\item {\bf Institutional review board (IRB) approvals or equivalent for research with human subjects}
    \item[] Question: Does the paper describe potential risks incurred by study participants, whether such risks were disclosed to the subjects, and whether Institutional Review Board (IRB) approvals (or an equivalent approval/review based on the requirements of your country or institution) were obtained?
    \item[] Answer: \answerNA{} 
    \item[] Justification: We did not utilize crowdsourcing or human subjects.

\item {\bf Declaration of LLM usage}
    \item[] Question: Does the paper describe the usage of LLMs if it is an important, original, or non-standard component of the core methods in this research? Note that if the LLM is used only for writing, editing, or formatting purposes and does not impact the core methodology, scientific rigorousness, or originality of the research, declaration is not required.
    \item[] Answer: \answerNA{} 
    \item[] Justification: We did not include LLM for developing our core methods.

\end{enumerate}

\newpage
\appendix

\section{Algorithm details}
\label{s:repdetails}

\begin{algorithm}[h]
\caption{Adaptive Token Merging (AToM)}
\label{alg:atom}
\textbf{Input}: Initial set of all tokens \(T\); Set of prompt tokens \(P\);\\
Number of model layers \(L\); \\
Maximum number of tokens to merge \(r_{\max}\)\\
\textbf{Initialize}: \(T'_{\text{final}} \leftarrow T\);
\begin{algorithmic}[1]
\FOR{\(l \in \{1, 2, \dots, \text{L}\}\)}
    \STATE \(T_{\text{attn}} \leftarrow \text{MSA}(T'_{\text{final}})\) 
    \STATE \(T_{\text{eligible}} \leftarrow T_{\text{attn}} \setminus P\) 
    \STATE \(\delta \leftarrow \frac{r_{\max}}{\text{L} - 1}\)
    \STATE \(n' \leftarrow \min(\delta \times (l - 1), r_{\max})\) 
    \STATE \(T'_{\text{merged}} \leftarrow \text{Merge}(T_{\text{eligible}}, n')\) 
    \STATE \(T_{\text{concat}} \leftarrow \text{Concat}(T'_{\text{merged}}, P)\) 
    \STATE \(T'_{\text{final}} \leftarrow \text{MLP}(T_{\text{concat}}, l)\)
\ENDFOR
\STATE \textbf{return} \(T'_{\text{final}}\)
\end{algorithmic}
\end{algorithm}

\begin{algorithm}[ht]
\caption{Adaptive Layer Dropping (ALD)}
\label{alg:ald}
\textbf{Input}: Input tensor \(X\); Keep ratio of layer \(\theta_{t,l}\);\\
Number of layers \(L\); Minimum ratio \(\bar{\theta}\); \\
Decay rate \(\gamma\); Spatial threshold \(\tau\);\\
Adjustment factor \(\alpha\)
\begin{algorithmic}[1]
\FOR{\(l \in \{1, 2, \dots, L\}\)}
    \IF{\((n(l) - n'(l)) > \tau\)}
        \STATE \(\alpha(l) \leftarrow \alpha\)
    \ELSE
        \STATE \(\alpha(l) \leftarrow 1\)
    \ENDIF
    \STATE \(\theta_{t,l} \leftarrow \alpha(l) \times \left( (1 - \bar{\theta})\exp(-\gamma \cdot t) + \bar{\theta} \right)\)
    \IF{\(\text{Bernoulli}(\theta_{t,l}) = 1\)}
        \STATE \(X_{\text{out}} \leftarrow \text{Exec}(l, X_{\text{out}})\) 
    \ELSE
        \STATE \(X_{\text{out}} \leftarrow X_{\text{out}}\)
    \ENDIF
\ENDFOR
\STATE \textbf{return} \(X_{\text{out}}\)
\end{algorithmic}
\end{algorithm}

\section{Implementation details}
\label{appendix:details}

Unless otherwise stated, we use the term \textbf{\ours-L2P} to denote \ours applied to L2P with a ViT-L backbone. Similarly, the notations \textbf{L}, \textbf{B}, and \textbf{Ti} are used to generally denote \textbf{ViT-L},
\textbf{ViT-B}, and \textbf{ViT-Ti} backbones, respectively, in the prompt update the stage of any prompt-based CL method.

\subsection{Prompt-based methods for \ours}

We present details on the multiple prompt-based rehearsal-free CL methods to which \ours is
integrated for enhancing resource efficiency.

\parlabel{L2P} positions prompts at the first layer of the transformer architecture~\cite{l2p}. These prompts are learnable parameters that dynamically evolve with the training process. The mechanism begins with a prompt pool \(P = \{p_1, p_2, ..., p_m\} \subset \mathbb{R}^{L_p \times D}\).

For a given input \(x_i^j\) in task \(T_i\), L2P computes a query feature \(q(x_i^j) \in \mathbb{R}^D\) to select the corresponding prompt. The prompt \(p^*\) is selected based on maximizing the cosine similarity with respect to the query \ie input data:
\begin{align}
p^* = \underset{p_k \in P}{\mathrm{argmax}}\sum_{c=1}^{L_p} \frac{\big\langle q(x_i^j), [p_k]_{(c,:)}^\top\big\rangle}{\|q(x_i^j)\|\|[p_k]_{(c,:)}^\top\|},
\end{align}
where $[p_k]_{(c,:)}$ is the $c$-th row of $p_k$.

The selected prompt \(p^*\) is concatenated with the input embedding \(z(x_i^j)\) to form the prompt-augmented input \(z'(x_i^j) = [p^*; z(x_i^j)]\). The training objective of L2P balances classification loss \(L_\text{class}\) and prompt-adjustment loss \(L_\text{prompt}\):
\begin{align}
L = L_\text{class}(z'(x_i^j), y_i^j) + L_\text{prompt}(p^*, q(x_i^j)).
\end{align}

\parlabel{DualPrompt} leverages prompts at multiple layers of the transformer architecture~\cite{dualp}. It introduces a general prompt \( g \in \mathbb{R}^{L_g \times D} \) and a set of task-specific prompts \( E = \{e_1, e_2, ..., e_T\} \subset \mathbb{R}^{L_e \times D}\). These prompts are incorporated at specified layers in the transformer model.

For a given input \(x_i^j\) in task \(T_i\), the model's transformer layers \(f\) are modified by attaching \(g\) and \(e_i\) to the layers, resulting in a prompted architecture \(f_{g,e_i}\). The feature transformation \(h_i^j\) for the input sample \(x_i^j\) is then obtained as:
\begin{align}
h_i^j = f_{g,e_i}(x_i^j).
\end{align}
Similar to L2P, DualPrompt optimizes \(L_\text{class}\) and \(L_\text{prompt}\):
\begin{align}
L = L_\text{class}(h_i^j, y_i^j) + L_\text{prompt}(g, e_i).
\end{align}

\parlabel{CODA-Prompt} decomposes learnable prompts into components and uses an attention mechanism from a pre-trained ViT model to select relevant prompts~\cite{codap}. Instead of a single prompt, CODA-Prompt learns a set of prompt components \( P = \{P_1, P_2, ..., P_M\} \). The final prompt \( p \) is calculated as a weighted sum:
\begin{align}
p = \sum_{m} \alpha_m P_m,
\end{align}
where weights \( \alpha \) are determined based on the query \( q(x) \) and keys \( K \in \mathbb{R}^{D \times M} \):
\begin{align}
\alpha = \gamma(q(x), K).
\end{align}
When the task changes, the current components are frozen, and a new one is added, promoting orthogonality, denoted as:
\begin{align}
L_{\text{ortho}}(B) = \|BB^\top - I\|^2,
\end{align}
where \( B \) represents \( P \), \( K \), or \( A \) (\( A \) represents the \textit{attention vector}). 
The full optimization target is:
\begin{align}
\min_{P^n, K^n, A^n, \phi^n} & \, L\left(f_\phi(f_{\theta, P, K, A}(x)), y\right) + \lambda \left( L_{\text{ortho}}(P) + L_{\text{ortho}}(K) + L_{\text{ortho}}(A) \right)
\end{align}
where \( P^n \), \( K^n \), and \( A^n \) are new components, and \( \lambda \) balances the orthogonality loss~\cite{codap}.

\parlabel{HiDe-Prompt} decomposes the CL objective into three parts—within-task prediction (WTP), task-identity inference (TII), and task-adaptive prediction (TAP)~\cite{hidep}. It manages uninstructed/instructed representations by modeling each class \(c\) with approximate distributions (\eg, Gaussian means \(\boldsymbol{\mu}_c\)). The overall combined loss is composed with WTP loss (\(\mathcal{L}_\mathrm{WTP}\)), TII loss (\(\mathcal{L}_\mathrm{TII}\)), TAP loss (\(\mathcal{L}_\mathrm{TAP}\)) and a contrastive regularization loss (\(\mathcal{L}_\mathrm{CR}\)):
\begin{align}
\label{eq:hidep_loss}
\mathcal{L}_{\mathrm{HiDe}} 
&= \mathcal{L}_\mathrm{WTP}\bigl(\boldsymbol{p}_t, f_\theta\bigr)
+ \mathcal{L}_\mathrm{TII}(\omega; \widehat{\mathcal{G}}_c)
+ \mathcal{L}_\mathrm{TAP}(\psi; \mathcal{G}_c)
+ \lambda\,\mathcal{L}_\mathrm{CR}\bigl(\boldsymbol{p}_t,\{\boldsymbol{\mu}_c\}\bigr),
\end{align}
where \(\boldsymbol{p}_t\) is the current prompt, \(\omega,\psi\) are auxiliary heads, 
and \(\widehat{\mathcal{G}}_c,\mathcal{G}_c\) denote distributions of uninstructed/instructed 
representations for old classes. This hierarchical approach excels under self-supervised pre-training.

\parlabel{ConvPrompt} generates layer-wise prompts via convolution on a shared embedding matrix \(\mathrm{SE}\)~\cite{convp}. For a layer \(\ell\) and head \(h\), it produces 
\(M\) prompt components \(\mathrm{PC}^{K}_{\ell,h,m}\), \(\mathrm{PC}^{V}_{\ell,h,m}\) by convolving \(\mathrm{SE}^K_{\ell,h}\) and \(\mathrm{SE}^V_{\ell,h}\) with $m$th prompt generator \(G_{\ell,h,m}\). The final prompt is computed by weighting these components via a projection network and prompt keys:
\begin{align}
\label{eq:convprompt}
P^{(K)}_{\ell+1,h}
&=
\sum_{m=1}^M s_{\ell,m}\,\mathrm{PC}^K_{\ell,h,m}, 
\qquad
P^{(V)}_{\ell+1,h}
=
\sum_{m=1}^M s_{\ell,m}\,\mathrm{PC}^V_{\ell,h,m}.
\end{align}
Here, \(s_{\ell,m}\) are similarity scores derived from \(\mathrm{[CLS]}\) embeddings, 
enabling a \emph{dynamic} prompt composition with small overhead.

\parlabel{OVOR} employs a single prompt throughout all tasks, in contrast to methods that maintain a large prompt pool~\cite{ovor}. To further refine decision boundaries, 
it introduces virtual outlier regularization (\textsc{VOR}), which generates synthetic outliers \(\mathbf{v}\in\mathcal{D}_{\mathrm{Outlier}}\) to penalize overconfident predictions outside the training distribution. OVOR’s loss is described as:
\begin{align}
\label{eq:ovor_outlier}
\mathcal{L}_{\mathrm{VOR}} 
= 
\mathbb{E}_{\mathbf{x}\sim \mathcal{D}^t}\!\Bigl[
\mathcal{H}\bigl(\widehat{\mathbf{y}}^{(\mathbf{x})}, \mathbf{y}\bigr)
\Bigr]
\;+\;
\lambda\,\mathbb{E}_{\mathbf{v}\,\sim\,\mathcal{D}_{\mathrm{Outlier}}}\!\Bigl[
\mathrm{hinge}\!\bigl(\tau_{\mathrm{Outlier}} - E(\mathbf{v})\bigr)
\Bigr],
\end{align}
where \(\mathcal{D}^t\) is the training set for task \(t\), \(\mathcal{H}\) is a cross-entropy or similar classification loss, \(\widehat{\mathbf{y}}^{(\mathbf{x})}\) the model prediction, and \(\mathbf{y}\) the ground-truth label for \(\mathbf{x}\). The term \(E(\mathbf{v})\) is an energy-based confidence measure, while penalizing overconfidence for outliers. Despite using only one prompt for all tasks, OVOR reports strong results by restricting overconfident predictions and minimizing overhead.

\parlabel{LAE} reframes prompts as an instance of parameter-efficient fine-tuning (PEFT), incorporating additional modules (Adapters or LoRA) to mitigate forgetting~\cite{lae}. It accumulates multi-task knowledge through online/offline PET modules and ensembles them at inference. Concretely, it defines an online PET \(\boldsymbol{\theta}_\mathrm{pet}^\text{on}\) for new tasks and an offline PET \(\boldsymbol{\theta}_\mathrm{pet}^\text{off}\) for older tasks:
\begin{align}
\label{eq:lae_ema}
\boldsymbol{\theta}_\mathrm{pet}^{\text{off}}
\;\leftarrow\; \alpha\,\boldsymbol{\theta}_\mathrm{pet}^{\text{off}}
\;+\;
(1-\alpha)\,\boldsymbol{\theta}_\mathrm{pet}^{\text{on}},
\end{align}
where \(\alpha\) is close to 1 (EMA update). During inference, LAE combines the outputs of these two experts to better handle both novel and historical tasks.

\subsection{Hyperparameters and configurations}
\label{s:setupinfo}

For all prompt-based CL methods, we employ the ADAM optimizer~\cite{adam} with hyperparameters \(\beta_1 = 0.990\) and \(\beta_2 = 0.999\), following their original implementation. For prompt selection, we set the prompt pool size to 10 and the prompt length to 5, following the original implementations of L2P, DualPrompt, and HiDe-Prompt. For CODA-Prompt, we implement a cosine-decay learning rate strategy 
described in~\cite{codap}, while other prompt-based baselines use a fixed learning rate of $1.875 \times 10^{-3}$ by default. 

We consistently use the mini-batch size of 16 for all methods to maintain a uniform computational load for each training iteration. For fair comparisons, each method performs 1875, 1080, and 2583 iterations per task for Split CIFAR-100~\cite{cifar}, Split ImageNet-R~\cite{imagenet_r}, and Split PlantDisease~\cite{plantdisease}, respectively. This setup ensures that among prompt-based methods, iteration time itself correlates linearly with energy usage, as training wall-clock time similarly reflects energy consumption, as discussed in the main paper.

\section{Competing methods with original setups}
\label{s:sotaoriginal}

\begin{table}[h]
\centering
\caption{Results on Split CIFAR-100, with ViT-B as the backbone (following their original setups).}
\small{
\begin{tabular}{lccccc}
\toprule
Method &  Acc. (\(\uparrow\)) & Fgt. (\(\downarrow\)) & GPU time (s) & Mem. (GB) \\
\midrule
REP-L2P & 87.0\(\pm\)1.3 & 4.5\(\pm\)0.1 & 4,421 & 4.5 \\
\midrule
L2P-B & 84.4\(\pm\)0.7 & 7.4\(\pm\)0.4 & 2,803 & 14.2 \\
DualPrompt-B & 85.3\(\pm\)0.8 & 5.2\(\pm\)0.1 & 2,624 & 12.3 \\
CODA-Prompt-B & 86.1\(\pm\)0.7 & 1.7\(\pm\)0.3 & 3,227 & 18.8 \\
HiDe-Prompt-B & 92.2\(\pm\)0.4 & 3.2\(\pm\)0.4 & 20,169 & 5.3 \\
ConvPrompt-B & 88.6\(\pm\)0.6 & 3.4\(\pm\)0.4 & 6,544 & 14.8 \\
LAE-Prefix10-B & 84.8\(\pm\)0.6 & 4.9\(\pm\)0.8 & 1125 & 2.8 \\
OVOR-B & 85.3\(\pm\)0.6 & 4.8\(\pm\)0.8 & 2,637 & 8.4 \\
\bottomrule
\end{tabular}
}
\label{tab:originalsotacomparison_cifar}
\end{table}

\begin{table}[h]
\centering
\caption{Results on Split ImageNet-R, with ViT-B as the backbone (following their original setups).}
\small{
\begin{tabular}{lccccc}
\toprule
Method &  Acc. (\(\uparrow\)) & Fgt. (\(\downarrow\)) & GPU time (s) & Mem. (GB) \\
\midrule
\ours-L2P &  75.3\(\pm\)1.5 & 3.6\(\pm\)0.3 & 2,542 & 4.5 \\
\midrule
L2P-B & 59.7\(\pm\)0.8 & 9.7\(\pm\)0.5 & 1,610 & 14.2 \\
DualPrompt-B & 67.5\(\pm\)0.9 & 4.7\(\pm\)0.2 & 1,508 & 12.3 \\
CODA-Prompt-B & 70.5\(\pm\)0.7 & 1.6\(\pm\)0.1 & 1,854 & 18.8 \\
HiDe-Prompt-B & 75.0\(\pm\)0.5 & 2.4\(\pm\)0.6 & 11,588 & 5.3 \\
ConvPrompt-B & 76.3\(\pm\)0.5 & 3.6\(\pm\)0.5 & 3,760 & 14.8 \\
LAE-Prefix10-B & 70.8\(\pm\)0.8 & 8.8\(\pm\)0.9 & 646 & 14.8 \\
OVOR-B & 73.2\(\pm\)0.8 & 4.5\(\pm\)0.9 & 1,516 & 8.4 \\
\bottomrule
\end{tabular}
}
\label{tab:originalsotacomparison_imr}
\end{table}

For the evaluation of baseline methods in the main paper, we adhere to the hyperparameters specified in the original studies, except for the \emph{number of iterations per task} and \emph{batch size}. We adopt a smaller batch size to account for on-device memory limitations, which prevent the use of the original batch sizes. Nonetheless, we present the results of \ours-L2P, comparing with L2P, DualPrompt, CODA-Prompt, HiDe-Prompt, ConvPrompt, LAE, and OVOR using their original hyperparameters in~\Cref{tab:originalsotacomparison_cifar} and~\Cref{tab:originalsotacomparison_imr}, respectively. In these experiments, we measure wall-clock GPU time rather than iteration time to ensure a fair comparison across diverse original settings. The results demonstrate that although these methods are based on the ViT-B backbone, most of them consume significantly more memory than \ours-L2P, primarily due to their originally large batch sizes.

\section{Preliminary empirical studies}
\label{s:preliminary_empirical_study}

\begin{figure*}[ht]
    \centering
    \includegraphics[width=\textwidth]{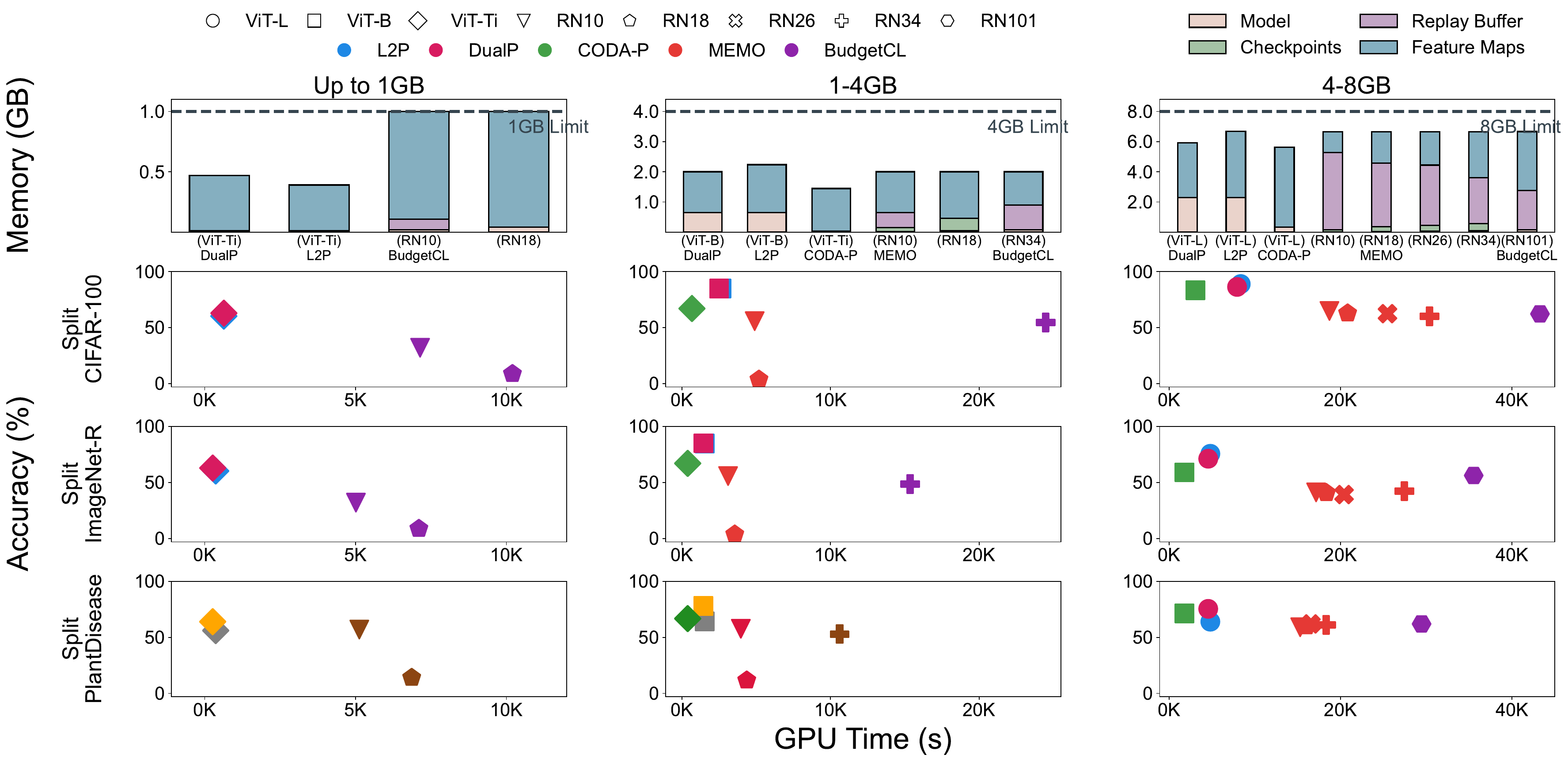}
    \vskip -0.05in
    \caption{Cost-accuracy trade-offs of various ViT- and CNN-based methods
    over three different memory budgets: up to 1GB, 1--4GB, and 4--8GB. The memory breakdown of each method is in the first row. Experiments on Split CIFAR-100, Split ImageNet-R, and Split PlantDisease are on the second, third, and fourth row, respectively. ViT-based methods consistently outperform CNN-based methods by a large margin.}
    \label{fig:energy_to_acc}
\end{figure*}

In this studies, we uncover the cost-accuracy trade-offs to highlight the impact of each baseline method on the accuracy gained relative to its training cost. Here, we limit the memory to a maximum of 8GB. We employ L2P~\cite{l2p}, DualPrompt~\cite{dualp}, and CODA-Prompt~\cite{codap} as representative ViT-based prompting methods. To demonstrate the effectiveness of prompting methods, we also incorporate two state-of-the-art CNN-based methods, such as BudgetCL~\cite{budgetcl} and MEMO~\cite{memo}. Both BudgetCL and MEMO improve model accuracy by leveraging spare memory to store and replay past samples, with MEMO additionally storing model checkpoints from history. These methods also use ImageNet pre-trained models as backbones: ResNet-10 (RN10; 5M), ResNet-18(RN18; 11M), ResNet-26 (RN26; 14M), ResNet-34 (RN34;22M), and ResNet-101 (RN101; 43M) models. 

For BudgetCL and MEMO, we adopt the SGD optimizer with a learning rate scheduling and a weight decay, following their original papers~\cite{budgetcl, memo}. Both methods construct mini-batches of 16 samples, with 8 samples from the new task and 8 from prior tasks in the replay buffer. We adopt the setup in~\cite{memo} and train for 200 epochs across all datasets. Thus, BudgetCL and MEMO perform more iterations per epoch with a larger replay buffer, resulting in longer training time and higher energy usage. For fair comparisons, we group methods by memory usage. 

\Cref{fig:energy_to_acc} visually represents the comparison results across device memory capacities and datasets. In each graph, the x-axis indicates GPU wall-clock time, which corresponds to energy cost (lower is better), while the y-axis indicates final average accuracy (higher is better). Thus, a more cost-effective method appears closer to the upper-left corner of the graph. Overall, ViT-based methods outperform CNN-based methods by a large margin, achieving 26--36\% higher accuracy with 45--90\% less time and energy spent under the same memory budget. We elaborate on two key characteristics that explain these results:

\parlabel{Scaling across varying backbone networks.\;} As we can see in~\cref{fig:energy_to_acc}, ViT-based methods scale well, with larger backbone networks overall yielding higher accuracy. In contrast, CNN-based methods scale poorly with increased backbone size. Specifically, on Split ImageNet-R, a more challenging dataset, MEMO/RN34 is only 3.0\% better than MEMO/RN18, despite being around twice larger. On Split CIFAR-100, increasing the backbone size yields only limited gains.

\parlabel{Memory efficiency.\;} CNN-based methods are often considered more memory-efficient because they use backbones a magnitude smaller than ViT models. Thus, it is commonly believed that ample memory can be allocated to replay samples or the past model checkpoints~\cite{memo}, which help improve model performance. However, in CL, which involves typical training iterations, a substantial amount of memory must be allocated for feature maps~\cite{zico}, leaving little room for memory buffers. For instance, for a device memory range of 1--4GB in~\Cref{tab:device_map}, the backbone of MEMO/RN18 consumes only 43MB of memory (6.5\% of ViT-L), but feature maps occupy as much as 1.7GB (more than ViT-based methods). This causes MEMO/RN18 to be short on memory for the replay buffer, dramatically degrading model accuracy, as observed in~\cref{fig:energy_to_acc}.

\begin{table}[h]
\caption{ViT- and CNN-based methods with specific backbone models are mapped
to the memory capacity they can fit into. DualP is DualPrompt, and CODA-P is CODA-Prompt.}
\centering
\footnotesize 
\setlength{\tabcolsep}{3pt}
\resizebox{0.75\textwidth}{!}{%
\begin{tabular}{cccc}
\toprule
Memory capacity & Edge device & ViT-based methods & CNN-based methods \\
\cmidrule(lr{5pt}){1-4}
4–8GB & Jetson Xavier NX &
L2P (ViT-L), DualP (ViT-L) &
MEMO (RN26, RN34) \\
& & CODA-P (ViT-B) & BudgetCL (RN101) \\
\cmidrule(l{5pt}r{5pt}){1-4}
1–4GB & Jetson Nano &
L2P (ViT-B), DualP (ViT-B) &
MEMO (RN10, RN18) \\
& & BudgetCL (RN34) & \\
\cmidrule(l{5pt}r{5pt}){1-4}
Up to 1GB & Raspberry Pi &
L2P (ViT-Ti), DualP (ViT-Ti) &
BudgetCL (RN10, RN18) \\
\bottomrule
\end{tabular}}%
\label{tab:device_map}
\vskip -0.1in
\end{table}

\section{Extended ablation studies}
\label{s:extend_ablation}

\begin{wraptable}{r}{0.48\linewidth}
\vspace{-6.6mm}
\centering
\caption{Component ablation for Split CIFAR-100. Ablating any component of \ours results in lower resource efficiency by increasing training time and memory consumption.}
\scriptsize
\setlength{\tabcolsep}{2pt}
\resizebox{\linewidth}{!}{
\begin{tabular}{lccccc}
\toprule
Ablated components & Acc. (\(\uparrow\)) & Fgt. (\(\downarrow\)) & Iter. time (ms) & Mem. (GB) \\
\midrule
\ours-L2P & 87.0\(\pm\)1.3 & 4.5\(\pm\)0.1 & 240 & 4.5 \\
\midrule
w/o (\atom + \ald) & 86.4\(\pm\)0.4 & 5.8\(\pm\)0.8 & 349 & 5.5 \\
w/o ($f_{\text{efficient}}$ + \ald) & 86.4\(\pm\)0.2 & 4.9\(\pm\)0.8 & 419 & 5.6 \\
w/o ($f_{\text{efficient}}$ + \atom) & 86.8\(\pm\)0.7 & 5.0\(\pm\)0.4 & 401 & 6.5 \\
w/o \ald & 85.9\(\pm\)0.5 & 5.0\(\pm\)0.7 & 270 & 4.5 \\
w/o \atom & 86.1\(\pm\)0.2 & 4.7\(\pm\)0.7 & 303 & 5.5 \\
w/o $f_{\text{efficient}}$ & 86.1\(\pm\)0.5 & 4.0\(\pm\)0.4 & 326 & 4.8 \\
\bottomrule
\end{tabular}
}
\label{tab:extended_component_ablation}
\end{wraptable}

As the paper mainly discusses the ablation study based on Split ImageNet-R, we here present an extended ablation studies of \ours using Split CIFAR-100 (10 tasks). The results are shown in~\Cref{tab:extended_component_ablation}. Each row reflects performance with the corresponding component removed. Similar to the findings from Split ImageNet-R, the ablation of any component within \ours for the Split CIFAR-100 dataset impacts resource efficiency. All components contribute to the reduction of computation time and memory usage, with \atom particularly standing out in both aspects of resource efficiency.

\begin{wraptable}{r}{0.48\linewidth}
\vspace{-2.7mm}
\centering
\caption{Algorithm validation for Split CIFAR-100. Changing our algorithms to conventional token merging (ToMe) or progressive layer dropping (PLD) harms model accuracy.}
\scriptsize
\setlength{\tabcolsep}{2pt}
\resizebox{\linewidth}{!}{
\begin{tabular}{lcccc}
\toprule
& Acc. (\(\uparrow\)) & Fgt. (\(\downarrow\)) & Iter. time (ms) & Mem. (GB) \\
\midrule
\ours-L2P & 87.0\(\pm\)1.3 & 4.5\(\pm\)0.1 & 240 & 4.5 \\
\midrule
w/ ToMe & 83.6\(\pm\)0.1 & 3.7\(\pm\)0.6 & 275 & 3.7 \\
w/ PLD & 84.4\(\pm\)0.4 & 4.9\(\pm\)0.8 & 259 & 4.5 \\
\bottomrule
\end{tabular}
}
\label{tab:extended_algorithm_validation}
\end{wraptable}

Next, we perform algorithm validation on Split CIFAR-100 (10 tasks), for which we present the results in~\Cref{tab:extended_algorithm_validation}. Consistent with the Split ImageNet-R results, using static token merging (ToMe)~\cite{token_merging} or progressive layer dropping (PLD)~\cite{progressive_drop} instead of adaptive token merging (\atom) or adaptive layer dropping (\ald) negatively affects model accuracy. This experiment further corroborates the efficacy of our proposed algorithms for optimizing resource usage without compromising the model accuracy in CL setup. We further extend our analysis from~\cref{tab:adjustratio} by evaluating two additional prompt-based CL methods, including HiDe-Prompt (denoted as \ours-HiDeP) and ConvPrompt (denoted as \ours-ConvP), with results detailed in~\cref{tab:extended_adjustratio}. These results show a trend consistent with~\cref{tab:adjustratio}.

\begin{table}[h]
\centering
\caption{REP over diverse \# of merged tokens (\(n\)) and \% of keep ratio (\(\theta\)), using HiDe-Prompt and ConvPrompt.}
\vskip 0.15in
\small{ 
{
    \begin{tabular}{lccccc}
    \toprule
     Method & $n$  (w/ \(\theta\)=0.5) & Acc. (\(\uparrow\)) & Fgt. (\(\downarrow\)) & Iter. time (ms) & Mem. (GB) \\
    \midrule
    \ours-HiDeP & 2 & 78.6\(\pm\)1.5 & 2.0\(\pm\)0.3 & 251 & 7.3\\
    \ours-HiDeP & 4 & 78.3\(\pm\)1.1 & 2.1\(\pm\)0.8 & 245 & 7.0\\
    \ours-HiDeP & 8 & 78.0\(\pm\)1.2 & 2.0\(\pm\)0.9 & 227 & 6.3\\
    \ours-ConvP & 2 & 79.1\(\pm\)1.0 & 3.4\(\pm\)0.5 & 460 & 5.5\\
    \ours-ConvP & 4 & 78.8\(\pm\)1.3 & 3.5\(\pm\)0.6 & 450 & 5.1\\
    \ours-ConvP & 8 & 78.5\(\pm\)0.8 & 3.7\(\pm\)1.1 & 417 & 4.1\\
    \midrule
    \midrule
    Method & \(\theta\)  (w/ $n$=8) & Acc. (\(\uparrow\)) & Fgt. (\(\downarrow\)) & Iter. time (ms) & Mem. (GB)\\
    \midrule
    \ours-HiDeP & 0.25 & 77.8\(\pm\)1.0 & 2.4\(\pm\)0.8 & 211 & 6.3\\
    \ours-HiDeP & 0.50 & 78.0\(\pm\)1.2 & 2.0\(\pm\)0.9 & 227 & 6.3\\
    \ours-HiDeP & 0.75 & 77.9\(\pm\)0.9 & 2.3\(\pm\)0.7 & 255 & 6.3\\
    \ours-ConvP & 0.25 & 78.2\(\pm\)1.1 & 3.9\(\pm\)0.4 & 387 & 4.1\\
    \ours-ConvP & 0.50 & 78.5\(\pm\)0.8 & 3.7\(\pm\)1.1 & 417 & 4.1\\
    \ours-ConvP & 0.75 & 78.3\(\pm\)0.4 & 4.0\(\pm\)1.2 & 469 & 4.1\\
    \bottomrule
    \end{tabular}}}
\label{tab:extended_adjustratio}
\vskip -0.1in
\end{table}

\section{Diverse task sequence lengths}
\label{s:task_scale}

We examine the scalability and robustness of our approach by employing two task sequence lengths, including 5 and 20 tasks, on the Split CIFAR-100 and Split ImageNet-R datasets, respectively. Longer task sequences simulate real-world scenarios more accurately, where CL methods require frequent model updates. The results are shown in~\Cref{tab:cifar_tasks} and~\Cref{tab:imr_tasks}, respectively. Across all scenarios, our method shows consistent preservation of both model accuracy and resource efficiency.

\begin{table*}[ht]
\centering
\caption{Results on Split CIFAR-100 organized as 5 tasks (5T) and 20 tasks (20T), respectively.}
\vskip 0.15in
\setlength{\tabcolsep}{3pt}
\resizebox{\textwidth}{!}{%
\begin{tabular}{l
                | c c  c c 
                | c c  c c
                | c c
                | c c}
\toprule
\textbf{Method} &
\multicolumn{4}{c}{\textbf{Accuracy (\(\uparrow\))}} &
\multicolumn{4}{c}{\textbf{Forgetting (\(\downarrow\))}} &
\multicolumn{2}{c}{\textbf{Iter. time (s)}} &
\multicolumn{2}{c}{\textbf{Mem. (GB)}} \\[3pt]
&
\multicolumn{2}{c}{Split CIFAR-100 (5T)} &
\multicolumn{2}{c}{Split CIFAR-100 (20T)} &
\multicolumn{2}{c}{Split CIFAR-100 (5T)} &
\multicolumn{2}{c}{Split CIFAR-100 (20T)} &
\multicolumn{2}{c}{} &
\multicolumn{2}{c}{} \\
\cmidrule(lr){2-5}\cmidrule(lr){6-9}\cmidrule(lr){10-11}\cmidrule(lr){12-13}
& 
w/o \ours & w/ \ours &
w/o \ours & w/ \ours &
w/o \ours & w/ \ours &
w/o \ours & w/ \ours &
w/o \ours & w/ \ours &
w/o \ours & w/ \ours \\
\cmidrule(lr){1-1}\cmidrule(lr){2-5}\cmidrule(lr){6-9}\cmidrule(lr){10-11}\cmidrule(lr){12-13}
L2P-L & 88.7\(\pm\)0.3 & 88.3\(\pm\)0.9 & 84.2\(\pm\)0.6 & 83.9\(\pm\)0.9 & 4.4\(\pm\)0.3 & 4.8\(\pm\)0.7 & 6.0\(\pm\)0.1 & 6.0\(\pm\)0.2 & 447 & 240 & 6.5 & 4.5 \\
DualPrompt-L & 87.9\(\pm\)0.5 & 87.2\(\pm\)0.9 & 83.7\(\pm\)0.7 & 83.0\(\pm\)0.9 & 4.4\(\pm\)0.2 & 4.7\(\pm\)0.8 & 6.3\(\pm\)0.6 & 6.8\(\pm\)0.9 & 424 & 208 & 5.9 & 4.3 \\
CODA-Prompt-L & 88.8\(\pm\)0.6 & 87.9\(\pm\)1.1 & 83.9\(\pm\)0.4 & 83.0\(\pm\)0.8 & 3.1\(\pm\)0.2 & 3.6\(\pm\)0.9 & 5.2\(\pm\)0.3 & 6.0\(\pm\)1.2 & 568 & 441 & 13.2 & 11.0 \\
HiDe-Prompt-L & 94.1\(\pm\)0.9 & 93.7\(\pm\)1.4 & 91.9\(\pm\)0.9 & 91.4\(\pm\)1.2 & 1.5\(\pm\)0.6 & 1.8\(\pm\)1.0 & 1.9\(\pm\)0.8 & 2.1\(\pm\)0.9 & 413 & 227 & 7.3 & 6.3 \\
ConvPrompt-L & 89.9\(\pm\)0.7 & 89.7\(\pm\)1.0 & 88.1\(\pm\)0.6 & 87.6\(\pm\)1.2 & 2.2\(\pm\)0.7 & 2.4\(\pm\)0.9 & 2.7\(\pm\)0.7 & 3.0\(\pm\)0.8 & 560 & 417 & 5.4 & 4.1 \\
LAE-Prefix10-L & 85.6\(\pm\)0.6 & 85.3\(\pm\)0.9 & 83.2\(\pm\)0.5 & 82.8\(\pm\)0.9 & 4.5\(\pm\)0.7 & 4.7\(\pm\)1.0 & 6.2\(\pm\)0.3 & 6.5\(\pm\)0.9 & 185 & 170 & 5.1 & 4.8 \\
OVOR-L & 86.9\(\pm\)0.8 & 86.4\(\pm\)1.1 & 85.4\(\pm\)0.6 & 85.0\(\pm\)0.9 & 3.2\(\pm\)0.7 & 3.6\(\pm\)1.0 & 3.0\(\pm\)0.4 & 3.2\(\pm\)0.9 & 467 & 362 & 7.5 & 6.8 \\
\bottomrule
\end{tabular}}%
\label{tab:cifar_tasks}
\vskip -0.1in
\end{table*}

\begin{table*}[ht]
\centering
\caption{Results on Split ImageNet-R organized as 5 tasks (5T) and 20 tasks (20T), respectively.}
\vskip 0.15in
\setlength{\tabcolsep}{3pt}
\resizebox{\textwidth}{!}{%
\begin{tabular}{l
                | c c  c c 
                | c c  c c
                | c c
                | c c}
\toprule
\textbf{Method} &
\multicolumn{4}{c}{\textbf{Accuracy (\(\uparrow\))}} &
\multicolumn{4}{c}{\textbf{Forgetting (\(\downarrow\))}} &
\multicolumn{2}{c}{\textbf{Iter. time (s)}} &
\multicolumn{2}{c}{\textbf{Mem. (GB)}} \\[3pt]
&
\multicolumn{2}{c}{Split ImageNet-R (5T)} &
\multicolumn{2}{c}{Split ImageNet-R (20T)} &
\multicolumn{2}{c}{Split ImageNet-R (5T)} &
\multicolumn{2}{c}{Split ImageNet-R (20T)} &
\multicolumn{2}{c}{} &
\multicolumn{2}{c}{} \\
\cmidrule(lr){2-5}\cmidrule(lr){6-9}\cmidrule(lr){10-11}\cmidrule(lr){12-13}
& 
w/o \ours & w/ \ours &
w/o \ours & w/ \ours &
w/o \ours & w/ \ours &
w/o \ours & w/ \ours &
w/o \ours & w/ \ours &
w/o \ours & w/ \ours \\
\cmidrule(lr){1-1}\cmidrule(lr){2-5}\cmidrule(lr){6-9}\cmidrule(lr){10-11}\cmidrule(lr){12-13}
L2P-L & 77.5\(\pm\)0.4 & 77.1\(\pm\)0.8 & 72.8\(\pm\)0.2 & 72.6\(\pm\)0.6 & 1.0\(\pm\)0.2 & 2.0\(\pm\)0.7 & 4.7\(\pm\)0.5 & 4.5\(\pm\)0.5 & 447 & 240 & 6.5 & 4.5 \\
DualPrompt-L & 73.1\(\pm\)0.5 & 72.5\(\pm\)0.9 & 69.5\(\pm\)0.4 & 69.0\(\pm\)1.4 & 2.2\(\pm\)0.5 & 2.6\(\pm\)1.3 & 5.0\(\pm\)0.9 & 5.8\(\pm\)0.9 & 424 & 208 & 5.9 & 4.3 \\
CODA-Prompt-L & 76.8\(\pm\)0.6 & 76.0\(\pm\)1.1 & 73.4\(\pm\)0.2 & 72.7\(\pm\)1.1 & 1.1\(\pm\)0.1 & 1.7\(\pm\)0.9 & 5.2\(\pm\)0.3 & 5.9\(\pm\)0.9 & 568 & 441 & 13.2 & 11.0 \\
HiDe-Prompt-L & 79.1\(\pm\)0.5 & 78.6\(\pm\)1.2 & 78.0\(\pm\)0.6 & 77.5\(\pm\)1.3 & 1.7\(\pm\)0.5 & 2.1\(\pm\)0.9 & 2.3\(\pm\)0.6 & 2.6\(\pm\)1.1 & 413 & 227 & 7.3 & 6.3 \\
ConvPrompt-L & 79.9\(\pm\)0.8 & 79.5\(\pm\)1.0 & 78.3\(\pm\)0.5 & 77.8\(\pm\)1.1 & 2.7\(\pm\)0.6 & 3.0\(\pm\)0.9 & 2.9\(\pm\)0.9 & 3.3\(\pm\)0.9 & 560 & 417 & 5.4 & 4.1 \\
LAE-Prefix10-L & 71.7\(\pm\)0.8 & 71.4\(\pm\)1.3 & 69.8\(\pm\)0.7 & 69.3\(\pm\)0.9 & 4.1\(\pm\)0.3 & 4.3\(\pm\)0.9 & 5.0\(\pm\)0.4 & 5.4\(\pm\)0.9 & 185 & 170 & 5.1 & 4.8 \\
OVOR-L & 76.7\(\pm\)0.8 & 75.9\(\pm\)1.3 & 74.6\(\pm\)0.5 & 73.9\(\pm\)0.9 & 5.1\(\pm\)0.7 & 5.5\(\pm\)1.0 & 4.6\(\pm\)0.6 & 5.1\(\pm\)0.9 & 467 & 362 & 7.5 & 6.8 \\
\bottomrule
\end{tabular}}%
\label{tab:imr_tasks}
\end{table*}

\section{Additional datasets}
\label{s:newdatasets}

We also expand our experiments on \ours by incorporating an additional
dataset: Split CUB-200~\cite{CUB200}, which is designed for fine-grained
image classification. Split CUB-200 contains 5,994 bird images categorized
into 200 classes, which are split into 5 tasks, each having 40 classes. 
As shown in~\cref{tab:newdataset}, integrating \ours achieves resource efficiency by bringing significant reductions in both training time and memory usage.

\begin{table}[ht]
\centering
\caption{Results on Split CUB-200, using ViT-L as the backbone model.}
\scriptsize
\setlength{\tabcolsep}{3pt}
\resizebox{0.9\textwidth}{!}{%
\begin{tabular}{l
                | c c
                | c c
                | c c
                | c c}
\toprule
\textbf{Method} &
\multicolumn{2}{c}{\textbf{Acc. (\(\uparrow\))}} &
\multicolumn{2}{c}{\textbf{Fgt. (\(\downarrow\))}} &
\multicolumn{2}{c}{\textbf{Iter. time (ms)}} &
\multicolumn{2}{c}{\textbf{Mem. (GB)}} \\
\cmidrule(lr){2-3}\cmidrule(lr){4-5}\cmidrule(lr){6-7}\cmidrule(lr){8-9}
& w/o \ours & w/ \ours &
  w/o \ours & w/ \ours &
  w/o \ours & w/ \ours &
  w/o \ours & w/ \ours \\
\midrule
L2P-L           & 74.7\(\pm\)0.9 & 74.7\(\pm\)1.0 & 6.7\(\pm\)1.4 & 6.7\(\pm\)0.3 & 447(\(\times\)1.9)  & 240  & 6.5(\(\times\)1.4) & 4.5 \\
DualPrompt-L    & 72.4\(\pm\)0.9 & 72.1\(\pm\)0.9 & 7.8\(\pm\)0.3 & 7.9\(\pm\)0.8 & 424(\(\times\)2.0)  & 208  & 5.9(\(\times\)1.4) & 4.3 \\
CODA-Prompt-L   & 79.5\(\pm\)0.8 & 79.0\(\pm\)1.1 & 5.8\(\pm\)0.9 & 6.0\(\pm\)0.7 & 568(\(\times\)1.3)  & 441  & 13.2(\(\times\)1.2) & 11.0 \\
HiDe-Prompt-L   & 86.8\(\pm\)0.5 & 86.6\(\pm\)1.0 & 1.8\(\pm\)0.8 & 1.8\(\pm\)1.3 & 413(\(\times\)1.8)  & 227  & 7.3(\(\times\)1.2)  & 6.3 \\
ConvPrompt-L    & 82.7\(\pm\)0.9 & 82.2\(\pm\)0.9 & 5.3\(\pm\)0.8 & 5.7\(\pm\)0.9 & 560(\(\times\)1.3)  & 417  & 5.4(\(\times\)1.3) & 4.1 \\
LAE-Prefix10-L    & 73.4\(\pm\)0.7 & 73.0\(\pm\)1.2 & 8.5\(\pm\)0.8 & 8.8\(\pm\)0.7 & 185(\(\times\)1.1)  & 170  & 5.1(\(\times\)1.1) & 4.8 \\
OVOR-L          & 78.6\(\pm\)0.7 & 78.0           & 5.9\(\pm\)0.9 & 6.3\(\pm\)0.9 & 467(\(\times\)1.3)  & 362  & 7.5(\(\times\)1.1) & 6.8 \\
\bottomrule
\end{tabular}}%
\label{tab:newdataset}
\vskip -0.1in
\end{table}

\section{Supplementary evaluations}
\label{s:extend_additional}
\leavevmode

\begin{wraptable}{r}{0.48\linewidth}
\vspace{-6.6mm}
\centering
\caption{Effect of \ours with HiDe-Prompt under self-supervised pretraining paradigms.}
\setlength{\tabcolsep}{2pt}
\vskip 0.15in
\resizebox{\linewidth}{!}{
        \begin{tabular}{lcccc}
            \toprule
             Method & Acc. (\(\uparrow\)) & Fgt. (\(\downarrow\)) & Iter. time (ms) & Mem. (GB) \\
            \midrule
            HiDe-iBOT & 71.5\(\pm\)0.3 & 1.4\(\pm\)0.4 & 130 & 4.1 \\
            \ours-iBOT & 70.7\(\pm\)0.9 & 0.7\(\pm\)0.7 & 76 & 2.4 \\
            \midrule
            HiDe-DINO & 68.6\(\pm\)0.2 & 1.8\(\pm\)0.2 & 130 & 4.1 \\
            \ours-DINO & 68.0\(\pm\)0.8 & 0.7\(\pm\)0.7 & 76 & 2.4 \\
            \bottomrule
            \end{tabular}}
\label{tab:sslcomparison}
\vspace{-6.6mm}
\end{wraptable}

\parlabel{Self-supervised pre-training.\;}
We evaluate \ours with HiDe-Prompt~\cite{hidep} under two self-supervised pre-training paradigms, including iBOT21K~\cite{ibot} and DINO~\cite{dino}, using the ViT-B backbone on Split ImageNet-R (10 tasks). As shown in~\Cref{tab:sslcomparison}, \ours reduces training time and memory usage by 14--18\%, with only 0.6--0.8\% of accuracy drop while forgetting remains improved.

\begin{wraptable}{r}{0.48\linewidth}
\vspace{-6.6mm}
\centering
\caption{Comparison of \ours-L2P with F-OAL on Split CIFAR-100.}
\setlength{\tabcolsep}{2pt}
\vskip 0.15in
\resizebox{\linewidth}{!}{
        \begin{tabular}{lccccc}
            \toprule
             Method & Acc. (\(\uparrow\)) & Fgt. (\(\downarrow\)) & Iter. time (ms) & Mem. (GB) \\
            \midrule
            \ours-L2P & 87.0\(\pm\)1.3 & 4.5\(\pm\)0.1 & 240 & 4.5 \\
            F-OAL & 87.6\(\pm\)1.4 & 5.1\(\pm\)0.3 & 359 & 1.6 \\
            \bottomrule
            \end{tabular}}
\vskip -0.2in
\label{tab:foalcomparison}
\end{wraptable}

\parlabel{Comparison with analytic learning-based CL method.\;}
We evaluate F-OAL~\cite{foal} as the representative analytic learning-based CL method on Split CIFAR-100 (10 tasks). \Cref{tab:foalcomparison} demonstrates that \ours-L2P can achieve compatible accuracy, with shorter training time. Since F-OAL is an online method, a perfect 1:1 comparison is tricky. Nonetheless, these results suggest that REP can be competitive with F-OAL’s energy efficiency while maintaining similar accuracy. Each method has distinct advantages and disadvantages depending on the context.

\section{Extended main results including forgetting metric}
\label{s:fullcomparison}

We present the extended results in~\Cref {tab:comparison}, including the forgetting metric. The result is shown in~\Cref{tab:forgetting}.

\begin{table*}[ht]
\centering
\caption{Forgetting and computational cost of all competing methods on Split CIFAR-100, Split ImageNet-R, and Split PlantDisease datasets. We report forgetting, iteration time, and memory usage both without and with \ours. Iteration time and memory usage are also shown as absolute values, with their multiples indicating how many times larger than \ours.}
\vskip 0.15in
\setlength{\tabcolsep}{3pt}
\resizebox{\textwidth}{!}{%
\begin{tabular}{l l
                | c c  c c  c c
                | c c
                | c c}
\toprule
& &
\multicolumn{6}{c}{\textbf{Forgetting (\(\downarrow\))}} &
\multicolumn{2}{c}{\textbf{Iter. time (ms)}} &
\multicolumn{2}{c}{\textbf{Mem. (GB)}} \\
\textbf{Model} & \textbf{Method} &
\multicolumn{2}{c}{Split CIFAR-100} &
\multicolumn{2}{c}{Split ImageNet-R} &
\multicolumn{2}{c}{Split PlantDisease} \\
\cmidrule(lr){3-4}\cmidrule(lr){5-6}\cmidrule(lr){7-8}\cmidrule(lr){9-10}\cmidrule(lr){11-12}
& &
w/o \ours & w/ \ours &
w/o \ours & w/ \ours &
w/o \ours & w/ \ours &
w/o \ours & w/ \ours &
w/o \ours & w/ \ours \\
\cmidrule(lr){1-2}\cmidrule(lr){3-8}
\cmidrule(lr){9-10}\cmidrule(lr){11-12}
\multirow{8}{*}{ViT-L}
& L2P            & 4.7\(\pm\)0.1 & 4.5\(\pm\)0.1 & 2.8\(\pm\)0.2 & 3.4\(\pm\)0.7 & 13.2\(\pm\)3.8 & 8.7\(\pm\)2.1 & 447(\(\times\)1.9) & 240 & 6.5(\(\times\)1.4) & 4.5 \\
& DualPrompt     & 4.5\(\pm\)0.7 & 4.9\(\pm\)0.9 & 3.8\(\pm\)0.4 & 4.1\(\pm\)0.9 & 14.8\(\pm\)3.0 & 10.1\(\pm\)3.7 & 424(\(\times\)2.0) & 208 & 5.9(\(\times\)1.4) & 4.3 \\
& CODA-Prompt    & 4.3\(\pm\)0.1 & 4.9\(\pm\)0.8 & 3.7\(\pm\)0.8 & 4.3\(\pm\)1.8 & 14.9\(\pm\)2.9 & 15.1\(\pm\)3.0 & 568(\(\times\)1.3) & 441 & 13.2(\(\times\)1.2) & 11.0 \\
& HiDe-Prompt    & 1.7\(\pm\)0.6 & 1.8\(\pm\)1.1 & 2.0\(\pm\)0.5 & 2.0\(\pm\)0.9 & 1.0\(\pm\)0.8 & 1.1\(\pm\)1.3 & 413(\(\times\)1.8) & 227 & 7.3(\(\times\)1.2) & 6.3 \\
& ConvPrompt     & 3.0\(\pm\)0.7 & 3.1\(\pm\)0.9 & 3.4\(\pm\)0.8 & 3.7\(\pm\)1.1 & 1.2\(\pm\)0.5 & 1.2\(\pm\)1.0 & 560(\(\times\)1.3) & 417 & 5.4(\(\times\)1.3) & 4.1 \\
& LAE-Prefix-10  & 5.4\(\pm\)0.8 & 5.5\(\pm\)1.0 & 4.5\(\pm\)0.5 & 4.6\(\pm\)1.2 & 15.5\(\pm\)0.8 & 15.9\(\pm\)1.4 & 185(\(\times\)1.1) & 170 & 5.1(\(\times\)1.1) & 4.8 \\
& OVOR           & 2.9\(\pm\)0.8 & 3.1\(\pm\)1.4 & 4.6\(\pm\)0.9 & 4.9\(\pm\)1.8 & 14.5\(\pm\)2.5 & 14.9\(\pm\)3.1 & 467(\(\times\)1.3) & 362 & 7.5(\(\times\)1.1) & 6.8 \\
& UpperBound     & --            & --            & --            & --            & --            & --            & 427              & --  & 9.8              & -- \\
\midrule
\multirow{8}{*}{ViT-B}
& L2P            & 6.6\(\pm\)0.8 & 6.7\(\pm\)0.9 & 6.6\(\pm\)1.2 & 6.1\(\pm\)1.2 & 25.9\(\pm\)2.8 & 23.5\(\pm\)2.9 & 143(\(\times\)1.4) & 102 & 2.3(\(\times\)1.2) & 2.0 \\
& DualPrompt     & 5.6\(\pm\)0.3 & 5.6\(\pm\)0.7 & 3.7\(\pm\)0.3 & 4.3\(\pm\)1.1 & 10.5\(\pm\)1.9 & 10.9\(\pm\)2.0 & 133(\(\times\)1.3) & 101 & 2.1(\(\times\)1.2) & 1.8 \\
& CODA-Prompt    & 6.0\(\pm\)0.2 & 6.5\(\pm\)1.5 & 6.1\(\pm\)0.8 & 6.9\(\pm\)0.9 & 18.8\(\pm\)1.1 & 19.1\(\pm\)2.9 & 164(\(\times\)1.1) & 151 & 6.9(\(\times\)1.2) & 5.7 \\
& HiDe-Prompt    & 1.6\(\pm\)0.8 & 1.6\(\pm\)1.4 & 1.2\(\pm\)0.8 & 1.9\(\pm\)1.3 & 1.0\(\pm\)0.8 & 1.2\(\pm\)1.0 & 130(\(\times\)1.7) & 76  & 4.1(\(\times\)1.7) & 2.4 \\
& ConvPrompt     & 3.7\(\pm\)0.6 & 3.9\(\pm\)1.6 & 5.5\(\pm\)0.7 & 5.9\(\pm\)1.4 & 1.6\(\pm\)0.8 & 1.8\(\pm\)1.5 & 381(\(\times\)1.4) & 279 & 2.2(\(\times\)1.1) & 2.0 \\
& LAE-Prefix-10  & 6.8\(\pm\)0.9 & 5.9\(\pm\)0.7 & 9.3\(\pm\)0.9 & 9.2\(\pm\)1.0 & 16.9\(\pm\)0.8 & 16.9\(\pm\)1.2 &  57(\(\times\)1.1) & 50  & 1.9(\(\times\)1.1) & 1.8 \\
& OVOR           & 5.7\(\pm\)0.6 & 5.9\(\pm\)1.5 & 4.9\(\pm\)0.7 & 5.5\(\pm\)1.4 & 20.8\(\pm\)0.9 & 21.4\(\pm\)1.6 & 134(\(\times\)1.1) & 123 & 3.1(\(\times\)1.1) & 3.0 \\
& UpperBound     & --            & --            & --            & --            & --            & --            & 143              & --  & 3.2              & -- \\
\midrule
\multirow{8}{*}{ViT-Ti}
& L2P            & 16.3\(\pm\)0.9 & 16.8\(\pm\)1.1 & 9.9\(\pm\)0.9 & 10.5\(\pm\)1.4 & 31.2\(\pm\)3.1 & 29.9\(\pm\)4.1 & 34(\(\times\)1.1) & 30 & 0.5(\(\times\)1.1) & 0.5 \\
& DualPrompt     & 15.7\(\pm\)2.2 & 15.8\(\pm\)3.0 & 8.9\(\pm\)0.8 & 9.2\(\pm\)1.0 & 25.0\(\pm\)2.2 & 26.1\(\pm\)2.8 & 34(\(\times\)1.1) & 31 & 0.4(\(\times\)1.1) & 0.4 \\
& CODA-Prompt    & 12.9\(\pm\)1.6 & 13.2\(\pm\)2.3 & 13.4\(\pm\)0.7 & 14.0\(\pm\)0.8 & 23.9\(\pm\)3.3 & 24.0\(\pm\)3.8 & 36(\(\times\)1.2) & 30 & 2.8(\(\times\)1.1) & 2.8 \\
& HiDe-Prompt    & 6.8\(\pm\)0.7 & 6.6\(\pm\)1.3 & 2.2\(\pm\)0.5 & 1.9\(\pm\)1.3 & 1.0\(\pm\)0.8 & 1.0\(\pm\)0.4 & 23(\(\times\)1.2) & 20 & 0.5(\(\times\)1.1) & 0.4 \\
& ConvPrompt     & 10.3\(\pm\)0.7 & 11.1\(\pm\)1.8 & 14.1\(\pm\)0.8 & 15.0\(\pm\)1.3 & 1.5\(\pm\)0.9 & 1.0\(\pm\)0.7 & 85(\(\times\)1.9) & 45 & 0.5(\(\times\)1.1) & 0.5 \\
& LAE-Prefix-10  & 17.1\(\pm\)0.3 & 17.1\(\pm\)0.9 & 17.0\(\pm\)0.5 & 17.8\(\pm\)0.9 & 27.6\(\pm\)0.8 & 27.9\(\pm\)1.4 & 15(\(\times\)1.3) & 12 & 0.4(\(\times\)1.1) & 0.4 \\
& OVOR           & 12.0\(\pm\)1.5 & 12.9\(\pm\)2.7 & 12.8\(\pm\)0.8 & 13.1\(\pm\)1.5 & 19.7\(\pm\)3.1 & 18.8\(\pm\)3.9 & 35(\(\times\)1.1) & 32 & 1.0(\(\times\)1.1) & 0.9 \\
& UpperBound     & --            & --            & --            & --            & --            & --            & 42               & -- & 0.6               & -- \\
\bottomrule
\end{tabular}}%
\label{tab:forgetting}
\end{table*}

\section{Additional related works}
\label{s:additional_related}

\subsection{Online or few-shot CL}
In on-device CL, the memory capacity for storing data from new tasks may be limited. Thus, one might consider adopting online or few-shot CL methods to save memory costs, as they require fewer new-task samples to be maintained in memory.
Offline CL (our setting) typically yields better accuracy with higher resource usage. However, memory allocation for new samples is not a major constraint in offline CL. All prompt-based methods require 155--269MB to store new samples for both Split CIFAR-100 and Split ImageNet-R, accounting for only 3.4--5.9\% of \ours-L2P's memory usage. Moreover, modern DL frameworks can store data in storage and proactively prefetch upcoming mini-batches for training, virtually eliminating memory concerns for new samples.

\subsection{Advantages of \ours over NAS}
While NAS (Neural Architecture Search) can be integrated into \ours to drop some layers, it usually involves searching the model from scratch, which is time- and computation-consuming. In contrast, our method applies adaptive schedules directly
to an existing pre-trained model, significantly saving computation time. So, it can naturally adapt to device-internal resource conditions without relying on external server resources.

\section{Impact of varying hyperparameter values}
\label{s:varied_hyperparameter}

Guided by the hyperparameter tuning strategies used in DualPrompt~\cite{dualp} and CODA-Prompt~\cite{codap}, we fine-tuned key hyperparameters for our \ald approach using \ours-L2P. Our focus is primarily on optimizing the threshold parameter ($\alpha$) and the adjustment factor ($\tau$), which are crucial for \ald's adaptability and efficiency. We conducted a hyperparameter search using 5-fold cross-validation, for which we designate 20\% of our training dataset as a validation set. \Cref{tab:hyperparam} summarizes the results.

We further evaluate the impact of varying prompt-selection-related hyperparameters using \ours-L2P on subsets of 5-datasets dataset introduced in~\cite{l2p}. \Cref{tab:hyperparam2} shows that the primary factor driving computation costs is prompt length, as it introduces more learnable parameters (\eg, with a pool size of 10, increasing prompt length from 5 to 10 increases memory usage by 500 MB and training time by 14\%). In contrast, increasing the prompt pool size has almost no impact on training time and memory usage.

\begin{table}[ht]
\centering
\caption{Results of ALD's hyperparameter tuning via 5-fold cross-validation.}
\label{tab:hyperparam}
\small
\begin{tabular}{r|*{10}{c}}
\toprule
\multirow{2}{*}{$\alpha$} & \multicolumn{10}{c}{$\tau$} \\
\cmidrule(lr){2-11}
 & 6 & 8 & 10 & 12 & 14 & 15 & 16 & 17 & 18 & 20 \\
\midrule
0.6 & 74.32 & 74.62 & 74.75 & 74.45 & 74.87 & 74.99 & 74.93 & 75.12 & 74.82 & 74.82 \\
0.7 & 74.17 & 74.38 & 74.45 & 74.98 & 74.18 & 75.06 & 75.15 & 75.26 & 74.43 & 74.31 \\
0.8 & 74.48 & 74.50 & 74.41 & 74.36 & 74.87 & 75.29 & 74.91 & 75.20 & 74.99 & 74.39 \\
0.9 & 74.49 & 74.36 & 74.77 & 74.38 & 74.58 & 75.33 & 75.34 & 75.42 & 74.88 & 74.96 \\
0.95 & 74.45 & 74.76 & 74.39 & 74.54 & 74.69 & 75.39 & 75.32 & 75.38 & 74.76 & 75.13 \\
\bottomrule
\end{tabular}
\end{table}

\begin{table}[ht]
\centering
\small
\caption{Effect of prompt pool size on subsets of 5-datasets. The prompt length is 10 by default.}
\setlength{\tabcolsep}{2pt}
\begin{tabular}{lcccc}
\toprule
Pool size & Acc. (\(\uparrow\)) & Iter. time (ms) & Mem. (GB)\\
\midrule
10 & 62.73 & 240 & 4.5 \\
15 & 63.07 & 240 & 4.5 \\
20 & 63.40 & 240 & 4.5 \\
30 & 63.78 & 241 & 4.5 \\

\bottomrule
\end{tabular}
\label{tab:hyperparam2}
\end{table}

\section{REP under varying computation budgets}
\label{s:impact_budget}

We compare \ours-L2P across various computation budgets defined by the
number of iterations per task. The comparison results using Split
CIFAR-100 (10 tasks) are shown in~\Cref{tab:suppl_budget}. Evidently, the accuracy of
\ours-L2P is comparable to that of L2P-L across six computation budgets, with the
performance gap getting increasingly marginal at lower compute budgets. We
observe that the overall accuracy for both methods does not drop
significantly with reduced compute budgets. This is attributed to the fact
that Split CIFAR-100 is considered a less complex CL benchmark.

\begin{table}[ht]
\centering
\caption{Impact of \ours under various computation budgets (\ie, number of iterations per task).}
\small{
\begin{tabular}{lcccccc}
\toprule
Method & 313 & 625 & 938 & 1250 & 1563 & 1875 \\
\midrule
\ours-L2P & 84.1 & 85.5 & 86.6 & 86.7 & 86.6 & 86.9 \\
L2P-L & 84.4 & 86.3 & 87.3 & 87.8 & 87.8 & 88.2 \\
\bottomrule
\end{tabular}
}
\label{tab:suppl_budget}
\end{table}

\section{\ours on edge devices}
\label{appendix:repedgedevices}

Miro~\cite{miro} introduces a dynamic approach for fine-tuning key design parameters of rehearsal-based CL methods to achieve high model accuracy while simultaneously reducing energy costs, \ie, high \emph{cost-effectiveness}.
Miro's methodology centers on identifying the suitable memory size within the device's capacity to accommodate both old and new samples for training. We compare \ours-L2P with Miro directly on a reference edge device, NVIDIA Jetson TX2.
This device is equipped with a 256-core NVIDIA Pascal GPU, 8GB of RAM, and a 32GB eMMC 5.1 drive. The RAM is a single unified memory shared by the CPU and GPU. We report energy usage as joules (J) obtained by multiplying power by time.
Power usage is measured by reading the built-in sensor values provided by Jetson devices for the GPU, RAM, CPU, and I/O connection.

\subsection{Power usage breakdown}
\label{s:gpupowerusage}

\begin{figure}[h]
   \centering
   \includegraphics[width=0.5\linewidth]{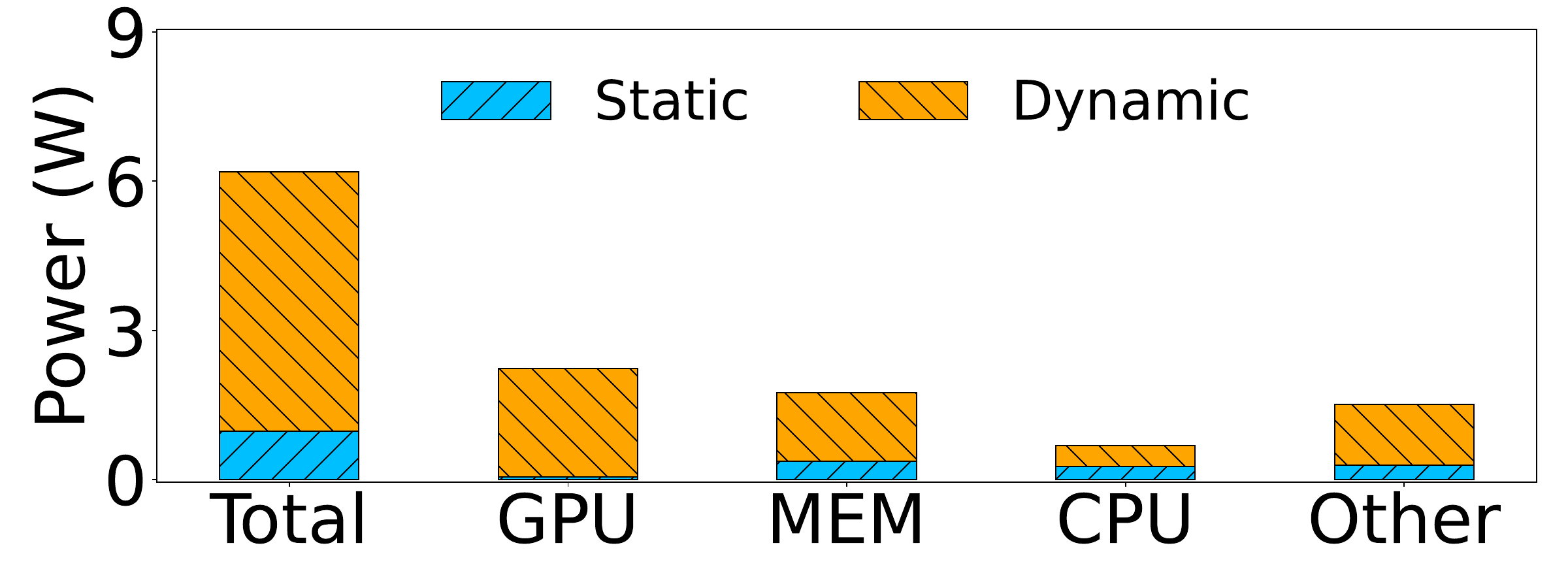}
   \caption{Power breakdown for training a ViT-B model on NVIDIA Jetson TX2.}
   \label{fig:power_consumptions}
\end{figure}

When implementing CL on edge devices, the majority of power consumption that influences energy usage during the training of a new task comes from GPU operations. \Cref{fig:power_consumptions} shows power consumption on NVIDIA Jetson TX2
across individual system components. Static power is measured when the system is inactive, while dynamic power is measured during the training of a ViT-B model. During the training process, power usage surges up to 6.5$\times$, with the GPU
contributing 60\% to the dynamic power.

\subsection{\ours vs Miro}
\label{s:repvsmiro}

\begin{figure}[h]
\centering
\includegraphics[width=0.5\linewidth]{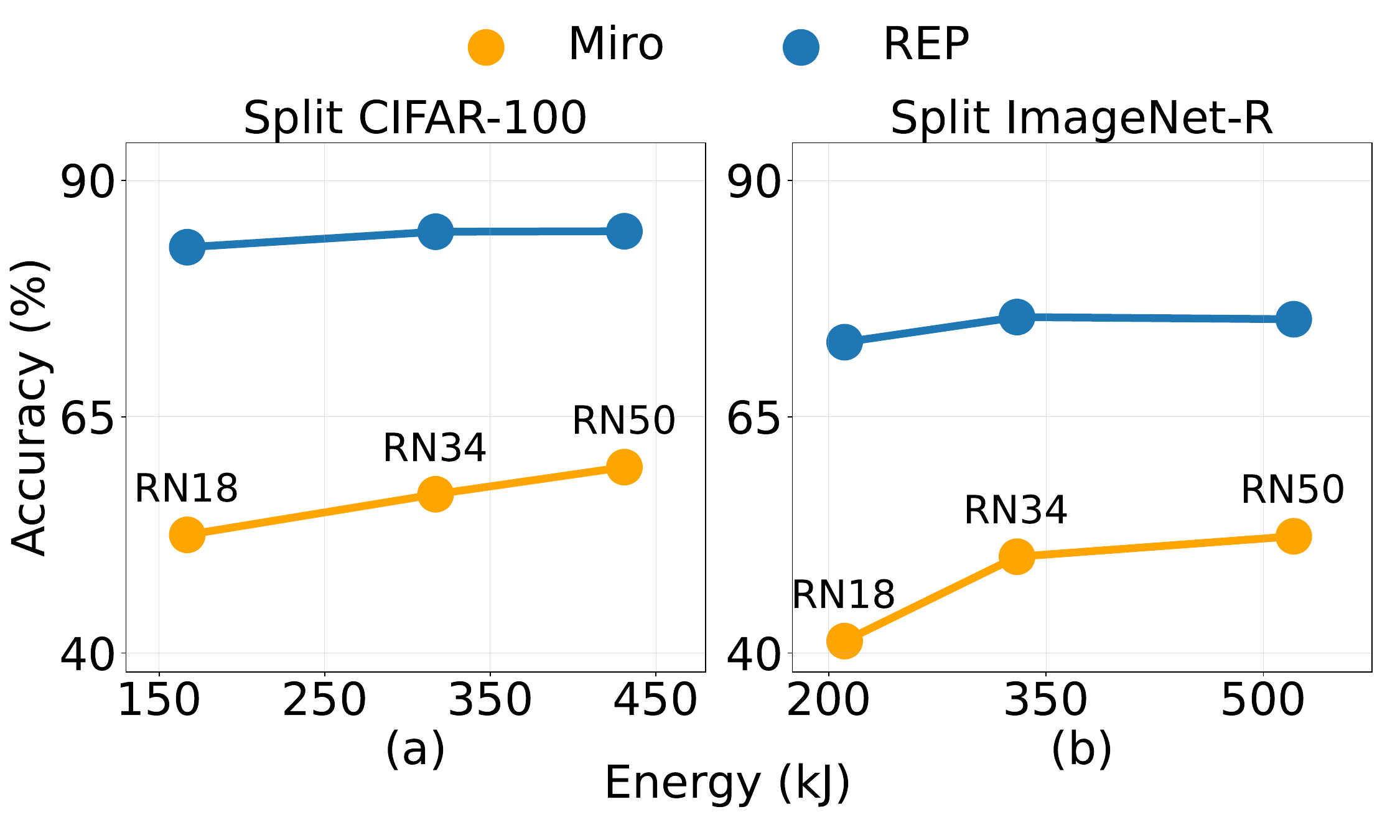}
\caption[]{Energy-accuracy trade-offs between \ours-L2P and Miro on Split CIFAR-100 (10 tasks) and Split ImageNet-R (10 tasks)
. To provide a spectrum of energy-accuracy trade-offs,
we use pre-trained ResNet-18 (RN18), ResNet-34 (RN34), and ResNet-50 (RN50)
for Miro.}
\label{fig:ours_miro}
\end{figure}

In this experiment, we explore energy-accuracy trade-offs of \ours-L2P and Miro. \Cref{fig:ours_miro} visually represents the comparison results. In each graph, the x-axis signifies total energy usage (lower is better), while the y-axis signifies final average accuracy (higher is better). Hence, a more cost-effective strategy is positioned closer to the upper-left corner of the graph. Regarding Miro, we maintain the hyperparameters as suggested in the original work~\cite{miro} but incorporate the use of pre-trained ResNet-18 (RN18; 11M), ResNet-34 (RN34; 22M), and ResNet-50 (RN50; 25M)\cite{resnet} models instead of non-pre-trained ones to enhance overall performance. For \ours-L2P, we vary the number of training iterations per task insertion to match the energy usage of Miro variants with different ResNet models.

When operating within the same energy budget, \ours-L2P consistently outperforms Miro variants, achieving 22--33\% higher accuracy across datasets. \ours-L2P demonstrates superior cost-effectiveness, especially on Split ImageNet-R (10 tasks) compared to Split CIFAR-100 (10 tasks). Both \ours-L2P and Miro prove to be memory-efficient to some extent as they fit comfortably within the on-device memory capacity. Although we explore larger ResNet models for Miro, we do not observe the accuracy levels comparable to \ours-L2P for either dataset. This experiment also underscores the importance of specifically optimizing vision transformers for on-device CL scenarios to advance performance boundaries.

\section{Societal impacts}
\label{s:societal}

The proposed method widely promotes AI for robotics agents and surveillance
systems with high efficiency.
Once the edge AI is easily deployable by the proposed method, the system may be used for monitoring
unwanted mass populations. Then, private information such as identity,
clothing information, and personal attributes (\eg, age, gender, etc.) could be
obtained by adversaries. Although our method has no intention to foster such
problematic cases, we do not currently propose secure solutions to prevent them from happening.

\end{document}